\documentclass[lettersize,journal]{IEEEtran}
\usepackage{cite}
\usepackage{amsmath,amssymb,amsfonts}
\usepackage{algorithmic}
\usepackage{graphicx}
\usepackage{textcomp}

\usepackage[ruled]{algorithm2e}
\usepackage[caption=false,font=footnotesize,labelfont=rm,textfont=rm]{subfig}
\usepackage{threeparttable}
\usepackage{multirow}
\usepackage[table]{xcolor}
\usepackage{enumitem}
\usepackage{makecell}
\usepackage{amssymb}

\newcommand{\etal}{{\em et al\,.}}       % et al.
\newcommand{\eg}{{\em e.g.}}           % e.g.
\newcommand{\ie}{{\em i.e.}}           % i.e.
\begin{document}

\title{Stitching, Fine-tuning, Re-training: \\ A SAM-enabled Framework for Semi-supervised 3D Medical Image Segmentation}
\author{Shumeng Li, Lei Qi, Qian Yu, Jing Huo, Yinghuan Shi$^*$, Yang Gao
\thanks{This work was supported by the NSFC Project (62222604, 62206052), China Postdoctoral Science Foundation (2024M750424), Fundamental Research Funds for the Central Universities (020214380120, 020214380128), State Key Laboratory Fund (ZZKT2024A14), the Postdoctoral Fellowship Program of CPSF (GZC20240252), Jiangsu Funding Program for Excellent Postdoctoral Talent (2024ZB242), and Jiangsu Science and Technology Major Project (BG2024031).}
\thanks{Shumeng Li, Jing Huo, Yinghuan Shi and Yang Gao are with the State Key Laboratory for Novel Software Technology, Nanjing University, China. They are also with National Institute of Healthcare Data Science, Nanjing University, China. (E-mail: lism@smail.nju.edu.cn, huojing@nju.edu.cn, syh@nju.edu.cn, gaoy@nju.edu.cn)}
\thanks{Lei Qi is with the School of Computer Science and Engineering, and the Key Lab of Computer Network and Information Integration (Ministry of Education), Southeast University, China. (E-mail: qilei@seu.edu.cn)}
\thanks{Qian Yu is with the School of Data and Computer Science, Shandong Women’s University, China. (E-mail: yuqian@sdwu.edu.cn)}
\thanks{The corresponding author of this work is Yinghuan Shi.}
}

\maketitle
\begin{abstract}
Segment Anything Model (SAM) fine-tuning has shown remarkable performance in medical image segmentation in a fully supervised manner, but requires precise annotations.
To reduce the annotation cost and maintain satisfactory performance, in this work, we leverage the capabilities of SAM for establishing semi-supervised medical image segmentation models. 
Rethinking the requirements of effectiveness, efficiency, and compatibility, we propose a three-stage framework, i.e., Stitching, Fine-tuning, and Re-training (SFR).
The current fine-tuning approaches mostly involve 2D slice-wise fine-tuning that disregards the contextual information between adjacent slices. 
Our stitching strategy mitigates the mismatch between natural and 3D medical images. 
The stitched images are then used for fine-tuning SAM, providing robust initialization of pseudo-labels.
Afterwards, we train a 3D semi-supervised segmentation model while maintaining the same parameter size as the conventional segmenter such as V-Net.
Our SFR framework is plug-and-play, and easily compatible with various popular semi-supervised methods.
We also develop an extended framework SFR$^+$ with selective fine-tuning and re-training through confidence estimation.
Extensive experiments validate that our SFR and SFR$^+$ achieve significant improvements in both moderate annotation and scarce annotation across five datasets. In particular, SFR framework improves the Dice score of Mean Teacher from 29.68\% to 74.40\% with only one labeled data of LA dataset.
The code is available at {\color{magenta}https://github.com/ShumengLI/SFR}.
\end{abstract}

% \vspace{10pt}

\section{Introduction}
\label{sec:intro}

\IEEEPARstart{R}{ecently}, general foundation models for visual segmentation~\cite{wang2023seggpt, zou2023segment, kirillov2023segment, khani2023slime} have attracted widespread attention in the field of medical images owing to their excellent segmentation and generalization capabilities.
Although these foundational models have made remarkable progress in medical image analysis, it is sometimes challenging to utilize a unified model to segment all medical images due to the inevitable factors, \eg, specific modalities, complex imaging techniques, and variable tissues. 
To tackle this issue, several recent works have been proposed to either focus on prompt engineering~\cite{gowda2024cc, xie2024sam, miao2024cross} or design adapters for fine-tuning~\cite{ma2023segment, wu2023medical, zhang2023customized, lin2023samus, cheng2024unleashing, chen2024ma, paranjape2024s} to borrow the ability of foundation model, \eg, MSA~\cite{wu2023medical} and SAMed~\cite{zhang2023customized} derived from SAM~\cite{kirillov2023segment}, to their specific tasks.

\begin{figure}[t]
   \centering
   \includegraphics[width=0.9\linewidth]{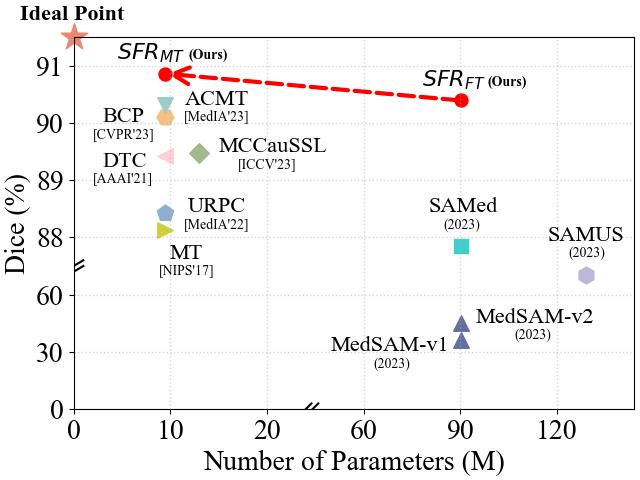}
   \caption{Comparison of our SFR framework with extended foundation models and semi-supervised medical segmentation methods on LA dataset~\cite{xiong2021global} with 16 labeled data.}
   \label{fig: intro}
\end{figure}

\IEEEpubidadjcol
We notice that most of these works~\cite{ma2023segment, wu2023medical, zhang2023customized, lin2023samus, chen2024ma, paranjape2024s, gowda2024cc} are fully supervised methods with fine-tuning or adaptation techniques employed.
However, fully supervised medical image segmentation relies on a great amount of precise annotations delineated by experienced experts, which makes the labeling process tedious, time-consuming, and even subjective. 

Recent trends~\cite{chen2023magicnet, xu2023ambiguity, bai2023bidirectional} have shown that in some cases, the performance of semi-supervised methods is almost comparable to that of fully supervised methods. 
For example, with 40\% annotation, \cite{chen2023magicnet} outperforms the fully supervised approach on BTCV dataset.
And the performance of~\cite{xu2023ambiguity} on LA dataset with 20\% labeled data is only 0.8\% lower than the fully supervised result. 
Therefore, we wonder, \textit{whether the current success of the foundation model could drive us to develop an effective and efficient model for semi-supervised medical image segmentation? }

With the aforementioned goal, we wish to revisit several important factors before designing our framework for semi-supervised medical image segmentation.

\textbf{How to initialize effectively?}
According to previous studies~~\cite{wu2023querying, yu2019uncertainty, tarvainen2017mean, ouali2020semi, chen2021semi, cai2023orthogonal, bai2023bidirectional, xu2023ambiguity, chen2023magicnet, miao2023caussl}, in the semi-supervised scenario, the quality of pseudo-labels in initialization stages plays an important role in the following segmentation. Unlike natural images, inter-slice continuity of 3D medical images is crucial for accurate target segmentation. Also, medical images usually have relatively low resolution. 
The existing strategies~\cite{ma2023segment, zhang2023customized, lin2023samus, cheng2024unleashing, chen2024aslseg} of fitting medical images, including directly enlarging 2D slices~\cite{ma2023segment} and resizing the positional embeddings~\cite{zhang2023customized, lin2023samus}, use slice by slice fine-tuning and disregard the inherent inter-slice correlation that exists in 3D images.
Therefore, is there a better way to boost the quality of initial pseudo-labels for medical images using a foundation model? 

\textbf{How to improve efficiency?} 
Foundation model is pre-trained on a large-scale dataset whose parameter size is relatively large. 
Existing fine-tuning methods~\cite{ma2023segment, wu2023medical, cheng2024unleashing, chen2024ma, paranjape2024s, gowda2024cc, xie2024sam, miao2024cross} still preserve the original parameter size as foundation model during inference or even introduce additional parameters.
During segmenting medical images, do we really need such a large parameter size? 
Firstly, the existing model only involving a small-size parameter indeed performs well in segmenting organs~\cite{bai2023bidirectional, miao2023caussl}. 
Secondly, we notice that very recent works~\cite{chen2022principle} depict the redundancy issue in the foundation model, revealing the large-scale pre-trained models are over-parameterized.
Thirdly, the appearance of medical images often has standardized views and relatively limited texture variants compared with natural images~\cite{alzubaidi2021novel}. 
Regarding these issues, is it possible to escape from over-parameterized foundation model while maintaining promising results?

\textbf{How to preserve compatibility?} 
On one hand, in recent years, semi-supervised learning has emerged as an appealing strategy and been widely applied to medical image segmentation tasks~\cite{jiao2023learning}, and a lot of semi-supervised medical image segmentation methods have been proposed. 
Could the foundation models be made to better serve existing semi-supervised methods?
On the other hand, the research progress in the field of computer vision and machine learning about semi-supervised learning still helps evolve new semi-supervised medical image segmentation models. 
Will our framework still be compatible with these new methods in the future? 

Being aware of these observations, we believe, that in the era of foundation model, a promising semi-supervised medical image segmentation framework should be performance effectiveness, parameter efficiency, and excellent compatableness. Thus, we propose a straightforward framework of \textbf{Stitching, Fine-tuning, and Re-training (SFR)} to accomplish our above goals.
We first develop a stitching strategy, performing a stitching operation on slices to produce an image that matches the high-resolution input, which better exploits inter-slice relationships and dimensional information. The stitched images are then fed into the SAM for fine-tuning.
Afterwards, we train a small-scale 3D segmentation model with the guidance of SAM while maintaining the same parameter size. The fine-tuned SAM provides a favorable initialization and is compatible with various 3D models. 
In addition, we develop the extended framework SFR$^+$, introducing confidence estimation and selective training strategy to enhance the utilization of unlabeled data.
We conduct the semi-supervised scenarios of \textit{moderate annotation}\footnote{Moderate refers to a commonly used level of annotation~\cite{xu2023ambiguity, chen2023magicnet}.} and \textit{scarce annotation}\footnote{Scarce means very few annotations such as 1 labeled data.} on five datasets. The experiments demonstrate our SFR and SFR$^+$ frameworks achieve extremely close performance to full supervision with moderate annotation and exhibit remarkable improvement with scarce annotation. As shown in Fig.~\ref{fig: intro}, our method in fine-tuning stage outperforms other extended foundation models and achieves a large improvement in re-training stage.

Our contribution could be summarized as follows:
\begin{itemize}
    \item We propose a novel framework leveraging the ability of the foundation model while ensuring performance under semi-supervised segmentation and further reducing labeling costs, which involves three stages, \ie, stitching, fine-tuning, and re-training.
    \item Our stitching strategy is simple yet effective in pseudo-labeling initialization, and largely distinct from current resize/directly fine-tuning strategy.
    \item Our parameter size during inference maintains the same level as the mainstream segmenter, \eg, V-Net~\cite{milletari2016v}, which is greatly smaller than that of foundation model. 
    \item Our framework is plug-and-play, which could be easily married to most existing popular semi-supervised segmentation methods.
\end{itemize}

\section{Related Work}
\label{sec:related work}

\begin{figure*}[t]
  \centering
   \includegraphics[width = 0.98\textwidth]{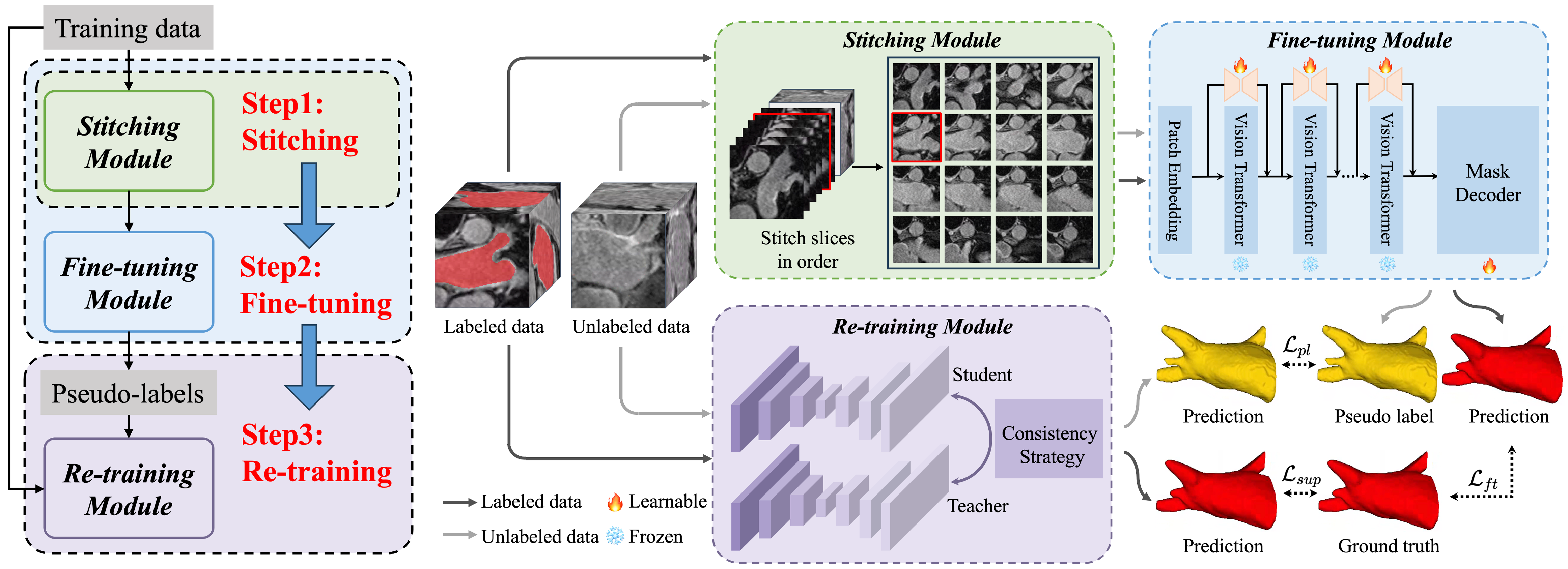}
   \caption{Overview of the proposed SFR framework, which includes three modules: Stitching, Fine-tuning and Re-training.}
   \label{fig: method}
\end{figure*}

%%%%%%%%%%%%%%%%%%%%% subsection %%%%%%%%%%%%%%%%%%%%%%%%
\subsection{Foundation Models in Medical Images}
%%%%%%%%%%%%%%%%%%% subsubsection %%%%%%%%%%%%%%%%%%%%%%
\subsubsection{Visual Foundation Models}
Nowadays, visual foundation models have gained significant attention and have shown impressive performance in various computer vision tasks including segmentation. 
Prominent examples of these models include SAM~\cite{kirillov2023segment}, SegGPT~\cite{wang2023seggpt}, SEEM~\cite{zou2023segment}, SLiMe~\cite{khani2023slime}, and SAM 2~\cite{ravi2024sam}, along with their extended applications~\cite{liu2023samm, wald2023sam, zhu2024medical}.
These models leverage large-scale image datasets to learn universal visual representations and demonstrate remarkable generalization ability.

In the field of medical images, UniverSeg~\cite{butoi2023universeg} achieves universal segmentation for 2D medical images by providing an example set of image-label pairs. STU-Net~\cite{huang2023stunet} is a foundation model specializing in CT modalities, with its largest variant consisting of 1.4 billion parameters. 
Furthermore, SAM~\cite{kirillov2023segment} has emerged as one of the most prevailing models for image segmentation, and many works extend it to medical images. SAM-Med2D~\cite{Cheng2023sammed2d} is a 2D model that fine-tunes SAM on 4.6 million medical images. SAM-Med3D~\cite{wang2023sammed3d} adopts a SAM-like architecture, but it is trained from scratch without utilizing the pre-trained weights of SAM.
Due to SAM's impressive performance and broad applicability, it serves as the default foundation model in our proposed framework. 

%%%%%%%%%%%%%%%%%%% subsubsection %%%%%%%%%%%%%%%%%%%%%%
\subsubsection{Adapt SAM to 3D Medical Images}
SAM's zero-shot capability is insufficient to ensure direct application in medical images~\cite{huang2023segment, mazurowski2023segment}. 
To extend the powerful segmentation ability to medical images, many works are devoted to fine-tuning with different image processing and fine-tuning strategies. 

For image processing, the disparity in image resolution between 3D medical images and pre-trained natural images poses a challenge, and two strategies have been proposed to tackle this issue. The first is upsampling fine-tuning~\cite{ma2023segment}, which involves upsampling each slice to match the input resolution directly. The second is small-size fine-tuning~\cite{zhang2023customized, lin2023samus}, which reduces the input size through bilinear interpolation. 
However, both of these strategies are based on 2D inputs, and for 3D medical images, the predictions need to be generated by segmenting each slice.
3DSAM-Adapter~\cite{gong20233dsam} and SAM-Med3D~\cite{wang2023sammed3d} extend SAM to 3D architecture, whereas they increase additional training overhead with a large model size.
In contrast, our stitching strategy is designed to accommodate variations in image dimension and resolution, creating large-sized stitched images that capture spatial information across adjacent slices effectively.

The fine-tuning approaches include fine-tuning only sub-parts parameters and incorporating adapters.
MedSAM~\cite{ma2023segment} fine-tunes the mask decoder of SAM and freezes the encoders, whereas the performance shows a lag behind medical-specific models, particularly in terms of boundary area. MSA~\cite{wu2023medical} and SAMed~\cite{zhang2023customized} adopt the parameter-efficient fine-tuning techniques, using adapter and low-rank-based strategy (LoRA)~\cite{hu2022lora} strategies for fine-tuning.

%%%%%%%%%%%%%%%%%%%%% subsection %%%%%%%%%%%%%%%%%%%%%%%%
\subsection{Semi-supervised Medical Image Segmentation}
Since the pixel-wise annotations require tremendous delineation time, semi-supervised learning (SSL) for segmentation aims to reduce the annotation burden by leveraging a large number of unlabeled samples along with a limited number of labeled samples~\cite{yu2019uncertainty, tarvainen2017mean, ouali2020semi, chen2021semi, cai2023orthogonal, bai2023bidirectional, xu2023ambiguity, chen2023magicnet, miao2023caussl}. 
Semi-supervised segmentation methods mainly include pseudo-labeling and consistency regularization. 
Self-training~\cite{bai2017semi} gradually generates pseudo-labels for unlabeled data through an iterative process to be jointly trained with labeled data. 
Mean Teacher~\cite{tarvainen2017mean} is a classic consistency regularization-based method that effectively reduces over-adaptation.
Xu \etal~\cite{xu2023ambiguity} improves the classical MT model to an ambiguity-consensus mean teacher model, and  Chen \etal~\cite{chen2023magicnet} develops a data augmentation strategy based on partition-and-recovery $N^3$ cubes.
Existing SSL methods have demonstrated promising results with moderate annotation levels. 
On one hand, our framework ensures seamless integration with various SSL models. 
On the other hand, we also apply the framework to challenging scenarios, further reducing the annotation requirements.

Some of the latest works~\cite{chen2024aslseg, miao2024cross} explore incorporating SAM into semi-supervised training. For example, ASLseg~\cite{chen2024aslseg} couples SAM with a specific semi-supervised model to refine pseudo-labels, while CPC-SAM~\cite{miao2024cross} uses SAM prompts for cross-branch teaching.
However, these methods still rely on slice-based processing and preserve a high computational load during inference. Our method could help improve performance while reducing inference costs. It is worth mentioning that our SFR and SFR$^+$ frameworks are compatible with various semi-supervised segmentation methods for medical images.

\textbf{Remark. }
Our framework explores the potential of leveraging SAM's capabilities in SSL medical image segmentation models, improving accuracy while reducing annotation costs. 
It is worth mentioning that our framework is compatible with most SSL methods for medical images. 

\section{Method}
\label{sec:framework}

%%%%%%%%%%%%%%%%%%%%% subsection %%%%%%%%%%%%%%%%%%%%%%%%
\subsection{Notations and Framework Overview}
Formally, we now provide our notation used in this paper. Given $m$ labeled images and $n$ unlabeled images, the $i$-th ($1 \leq i \leq m$) labeled image and its ground truth are denoted as $\mathbf{X}^l_i$ and $\mathbf{Y}^l_i$. The $j$-th ($1 \leq j \leq n$) unlabeled image is denoted as $\mathbf{X}^u_j$. Here, $\mathbf{X}^l_i$, $\mathbf{X}^u_j \in \mathbb{R}^{H \times W \times D}$, $\mathbf{Y}^l_i \in \{0,1,...,K-1\}^{H \times W \times D}$, where $H$, $W$, $D$ indicate the corresponding dimensionality of 3D medical images and the input patch size for training is square, \ie, $H=W$. $K$ is the number of different classes to segment. 

As illustrated in Fig.~\ref{fig: method}, we build our SFR framework, which consists of the following three modules: the Stitching Module, the Fine-tuning Module, and the Re-training Module. The stitching module mitigates the mismatch between natural and 3D medical images. The stitched images are input into SAM to fine-tune and provide initial pseudo labels for the semi-supervised module. Afterwards, the 3D semi-supervised segmentation model is trained.
Furthermore, we enhance our framework by proposing SFR$^+$, which selectively fine-tunes and re-trains through confidence estimation.

$\blacktriangleright$ \underline{\textit{\textbf{Step 1: Stitching Module.}}}
As aforementioned, by reducing the resolution difference between pre-trained samples (\ie, natural image) and fine-tuned samples (\ie, medical image), the stitching module could transform a 3D labeled volume $\mathbf{X}^l_i$ to a large-sized 2D image $\mathbf{M}^l_i \in \mathbb{R}^{Hd\times Wd}$ with a slice stitching function $\texttt{F}_\texttt{C}(\cdot)$. Also, the stitched ground truth $\mathbf{N}^l_i$ is obtained similarly. $\mathbf{M}^l_i$ and $\mathbf{N}^l_i$ are arranged in a $d\times d$ grid with $d = \lceil \sqrt{D} \rceil$, and we stitch zeros after all slices if $d\times d > D$.
  \begin{equation}
   \mathbf{M}^l_i = \texttt{F}_\texttt{C}(\mathbf{X}^l_i), \qquad
   \mathbf{N}^l_i = \texttt{F}_\texttt{C}(\mathbf{Y}^l_i).
  \end{equation}

$\blacktriangleright$ \underline{\textit{\textbf{Step 2: Fine-tuning Module.}}} 
We first utilize the stitched labeled images $\mathbf{M}^l_i$ along with their ground truth $\mathbf{N}^l_i$ to fine-tune a SAM parameterized by $\theta$ via popularly used strategy, \eg, LoRA~\cite{hu2022lora, zhang2023customized}.
This aims to narrow the possible distribution shift between natural and medical images. Taking LoRA as an example, we denote the fine-tune function as $\texttt{F}_\texttt{LoRA}(\cdot)$, which is parameterized by $\theta$ with input as $\mathbf{M}^l_i$ and $\mathbf{N}^l_i$. The updated SAM with optimal parameter $\theta^*$ is obtained as follows:
  \begin{equation}
   \theta^* = \arg\min_{\theta} \sum_{i=1}^m\texttt{F}_\texttt{LoRA}(\mathbf{M}^l_i, \mathbf{N}^l_i; \theta). 
  \end{equation}

Then, we produce high-quality pseudo-labels for unlabeled images by using the prediction function $\texttt{F}_\texttt{FT}(\cdot)$ of fine-tuned SAM, and generate 3D pseudo-labels by a stitching inverse transform $\texttt{F}_\texttt{C}^\texttt{-1}(\cdot)$ as follows:
  \begin{equation}
   \hat{\mathbf{Y}}^u_j = \texttt{F}_\texttt{C}^\texttt{-1}\Bigl(\texttt{F}_\texttt{FT}\bigl(\texttt{F}_\texttt{C}(\mathbf{X}^u_j); \theta^*\bigr)\Bigr).
  \end{equation}

\begin{figure}[t]
  \centering
   \includegraphics[width=0.95\linewidth]{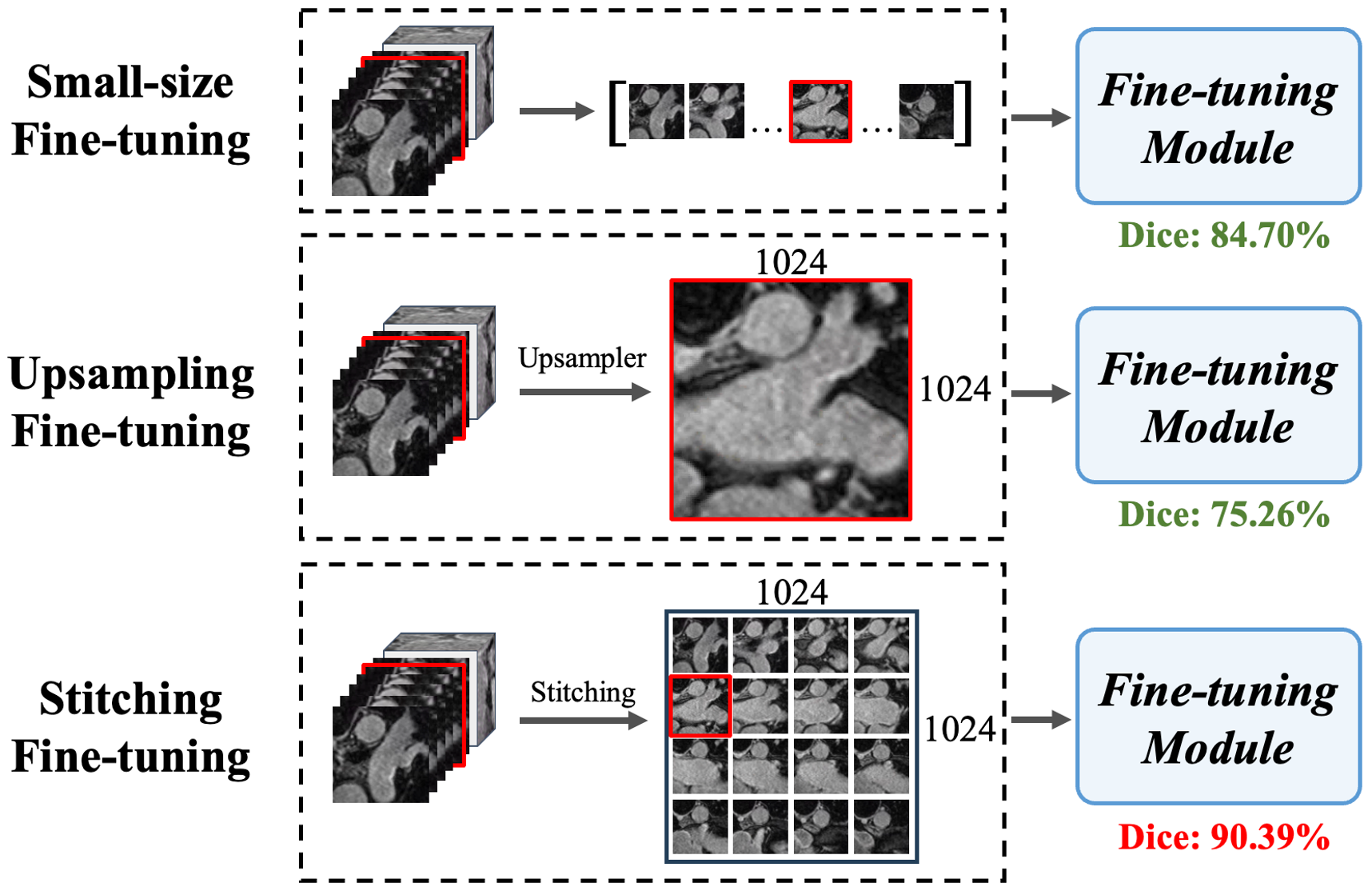}
   \caption{Comparison of different input strategies. Small-size fine-tuning reduces the input size through bilinear interpolation and upsampling fine-tuning directly upsamples each slice. Taking $d = 4$ as an example.}
   \label{fig: three_ft}
\end{figure}

$\blacktriangleright$ \underline{\textit{\textbf{Step 3: Re-training SSL Module.}}} 
The recent SSL network, such as the popular self-training, mean teacher, and the most advanced ACMT, MagicNet methods, could be the alternatives in this module. Specifically, the SSL network learns pseudo-labels from fine-tuned SAM. We denote the SSL network as $\texttt{F}_\texttt{S}(\cdot)$ parameterized by $\omega$ and the re-training module with optimal $\omega^*$ is as follows: 
  \begin{equation}
  \begin{aligned}
   \omega^* = \arg\min_{\omega} \bigl(\sum_{i=1}^{m}\texttt{F}_\texttt{S}(\mathbf{X}^l_i, \mathbf{Y}^l_i; \omega) + \lambda \sum_{j=1}^{n}\texttt{F}_\texttt{S}(\mathbf{X}^u_j, \hat{\mathbf{Y}}^u_j; \omega)\bigr), 
  \end{aligned}
  \end{equation}
where $\lambda$ acts as a tradeoff between two terms.

\subsection{SFR Framework}
\subsubsection{Stitching Module}
\label{sec:concat}

%%%%%%%%%%%%%%%%%%%%% subsection %%%%%%%%%%%%%%%%%%%%%%%%
To adapt from 2D natural images to 3D medical images, we recognize that the input resolution and image dimension are crucial factors. The inter-slice spatial information of 3D volumes is relevant for target recognition, and it is difficult for the large model trained at high-resolution images to generalize to low-resolution medical image slices.
Inspired by this observation, our stitching strategy, illustrated in Fig.~\ref{fig: three_ft}, matches medical images to natural image resolution, and supplements the spatial arrangement specific to 3D medical images.
The input spatial resolution of the pre-trained SAM model is $1024 \times 1024$. 
Our stitching strategy arranges the 3D volume (either a raw 3D image or a 3D patch) slice by slice into a $d \times d$ grid, producing a 2D image of size $1024 \times 1024$. 
Regarding the variability in slice sizes across different medical datasets, there is a performance trade-off between the slice resolution and the number of stitched slices. Our method could effectively manage images with different numbers of slices.
For the small-scale slices, such as the LA~\cite{xiong2021global} dataset, we use a raw 3D image as an input volume. For large-scale slices, such as the BTCV~\cite{landman2015miccai} dataset, we follow the common 3D processing way where dividing the volume into patches as an input volume, and then stitch all slices to achieve the final size of $1024\times 1024$, avoiding downsizing a large slice directly.
Compared to the small-size input fine-tuning method~\cite{zhang2023customized, lin2023samus} and the direct upsampling fine-tuning methods~\cite{ma2023segment} in Fig.~\ref{fig: three_ft}, we find our stitching strategy effectively addresses the challenges of image dimension and resolution differences. 

We thoroughly investigate the stitching strategy from \textit{slice continuity} and \textit{contextual integrity}.

\begin{figure}[t]
  \centering
   \includegraphics[width=0.95\linewidth]{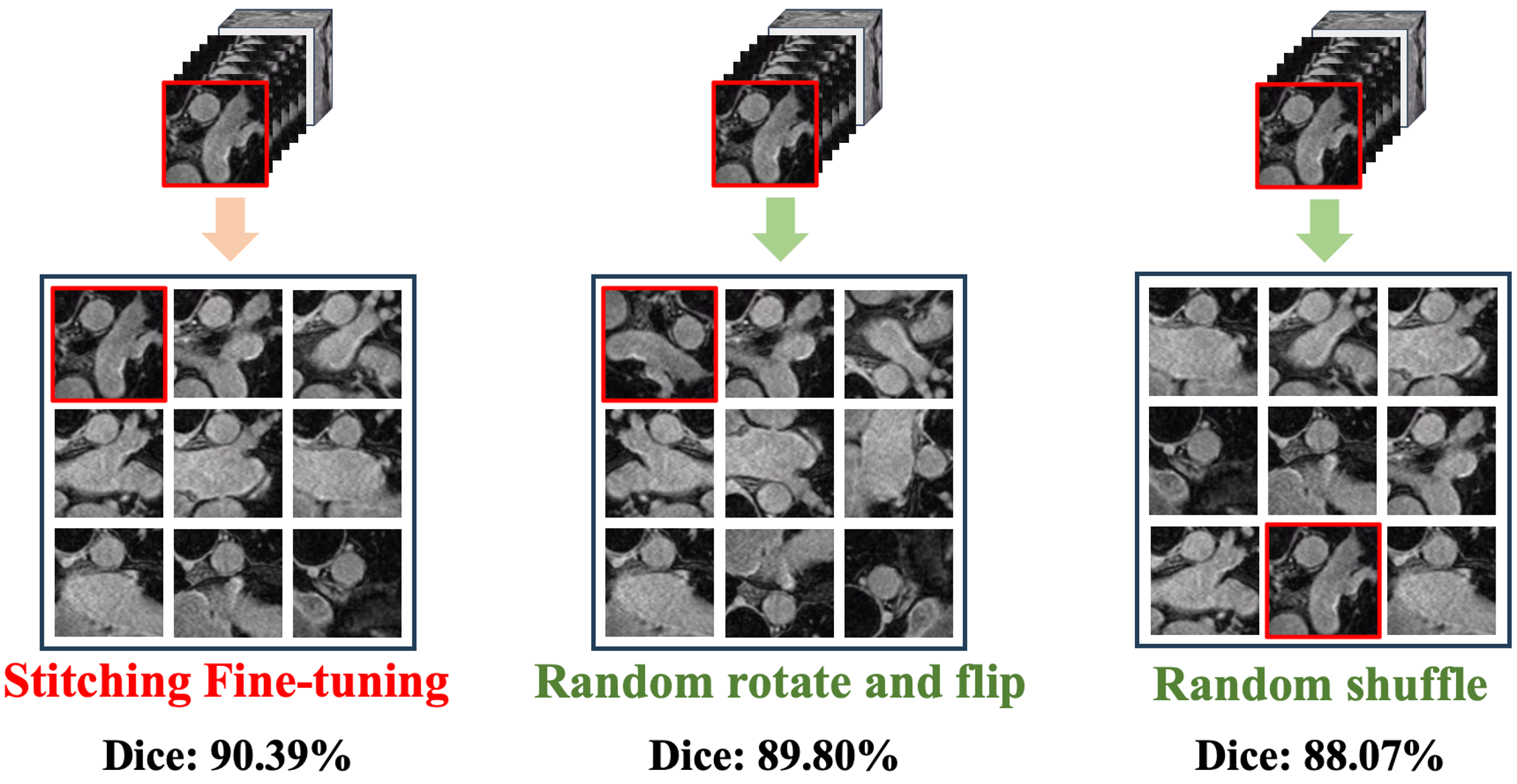}
   \caption{Disrupting slice continuity. Comparison with random slice rotation and flipping, and random shuffling. $d = 3$ as an example.}
   \label{fig: shuffle}
\end{figure}

\textbf{Slice Continuity.}
Due to the inherent spatial continuity of 3D medical images, we explore the relationship between slice stitching and slice order. For a stitched 2D image, the segmentation model learns the feature correlation across slices through the self-attention mechanism, so it struggles to capture contextual information and coherence of shape without the slice order. 
To investigate the importance of slice continuity, we disrupt the continuity in two ways:
1) Randomly shuffling the order of the slices;
2) Randomly rotate and flip each individual slice.
The same slicing operation is also applied to the ground truth masks to maintain consistency between the input data and labels.
As illustrated in Fig.~\ref{fig: shuffle}, it has been observed that disrupting the slice order leads to the loss of shape coherence and a decrease in performance, and the results emphasize the importance of slice continuity in 3D organ segmentation. 

\begin{figure}[t]
    \centering
    \includegraphics[width = 1.0\linewidth]{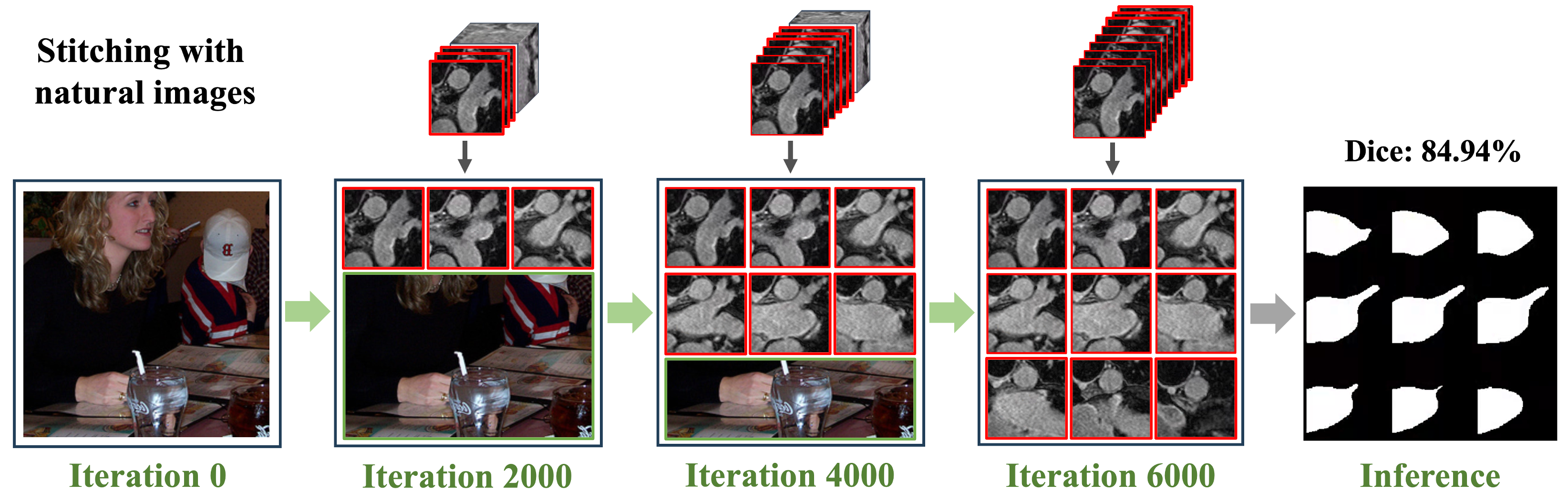}
    \caption{Disrupting contextual integrity. Comparison with stitching with natural images. $d=3$ as an example. }
    \label{fig:with_natural}
\end{figure}

\textbf{Contextual Integrity.}
Our stitching module reorganizes a volume (3D raw image or 3D patch) into a $1024 \times 1024$ image, enabling a complete representation of the volume within a single image.
To explore the impact of contextual integrity on stitching, we stitch the medical slices with natural images while keeping the resolution of each slice constant. As shown in Fig.~\ref{fig:with_natural}, we transition from natural images to medical images by progressively increasing the number of medical slices.
Specifically, at the beginning of the training process, we incorporate natural images from the PASCAL VOC 2012 dataset~\cite{pascal-voc-2012}, which contains a total of 2,913 images. For each iteration, we randomly select a batch of natural images from this dataset. As training progresses, we incrementally replace parts of the natural images with medical image slices, such that by the final stage, only medical slices are utilized. For testing, we consistently use fully stitched medical image slices.
Although this approach seems to gradually adjust from natural image features to medical image features, it actually disrupts the integrity of the anatomical structure in medical images and leads to a decrease in performance. 

These observation results reveal the importance of maintaining \textit{slice continuity} and \textit{contextual integrity} for our stitching strategy, effectively bridging the domain and spatial dimensional gaps between natural and medical images.

\subsubsection{Fine-tuning Module}
\label{sec:finetune}

Our fine-tuning module performs fine-tuning on the vision foundation model.
As one of the most popular universal image segmentation models, SAM serves as the default setting of our fine-tuning module, denoted as $\texttt{F}_\texttt{FT}(\cdot)$.
We strip away all the prompts and perform automatic segmentation during inference. Our framework is not limited to a specific fine-tuning strategy, which could be used in different strategies.
We uniformly denote the fine-tuning loss $\mathcal{L}_{ft}$: 
\begin{equation}
\label{eq: ft}
\mathcal{L}_{ft} = \frac{1}{2} \bigl(\mathcal{L}_{Dice} (\mathbf{P}^l_i, \mathbf{N}^l_i) + \mathcal{L}_{ce}(\mathbf{P}^l_i, \mathbf{N}^l_i)\bigr), 
\end{equation}
where $\mathbf{P}^l_i=\texttt{F}_\texttt{FT}(\mathbf{M}^l_i)$ is the prediction of fine-tuning module. 

The previous studies mainly involve fine-tuning only sub-parts parameters~\cite{ma2023segment} and incorporating adapters~\cite{wu2023medical, zhang2023customized}. 

\textbf{Sub-parts Fine-tuning. }
Sub-parts fine-tuning methods directly modify the model parameters. 
MedSAM-v1~\cite{ma2023segment} freezes the image encoder and prompt encoder by only fine-tuning the mask decoder, and MedSAM-v2 fine-tunes both image encoder and mask decoder. However, the overall performance still lags behind expert models for medical image segmentation, particularly in terms of boundary consensus.

{\textbf{Adapter Tuning. }
Adapter tuning~\cite{houlsby2019parameter, wu2023medical} is to insert adapters into the original fundamental model, and only tune adapters while leaving all pre-trained parameters frozen. An adapter consists of a down-projection, ReLU activation, and up-projection layers. 
The low-rank-based fine-tuning strategy (LoRA)~\cite{hu2022lora} injects trainable low-rank decomposition matrices into the layers of the pre-trained model. SAMed~\cite{zhang2023customized} freeze the image encoder of SAM, adopt LoRA by adding a bypass, and fine-tune the mask decoder. 

Since LoRA can be merged with the original pre-trained weights for inference, we adopt it as our fine-tuning module method $\texttt{F}_\texttt{LoRA}(\cdot)$. 
Following \cite{zhang2023customized}, for the classification head of SAM, ambiguity prediction is replaced by the determined prediction output for each semantic category.

\begin{figure}[t]
  \centering
   \includegraphics[width=0.95\linewidth]{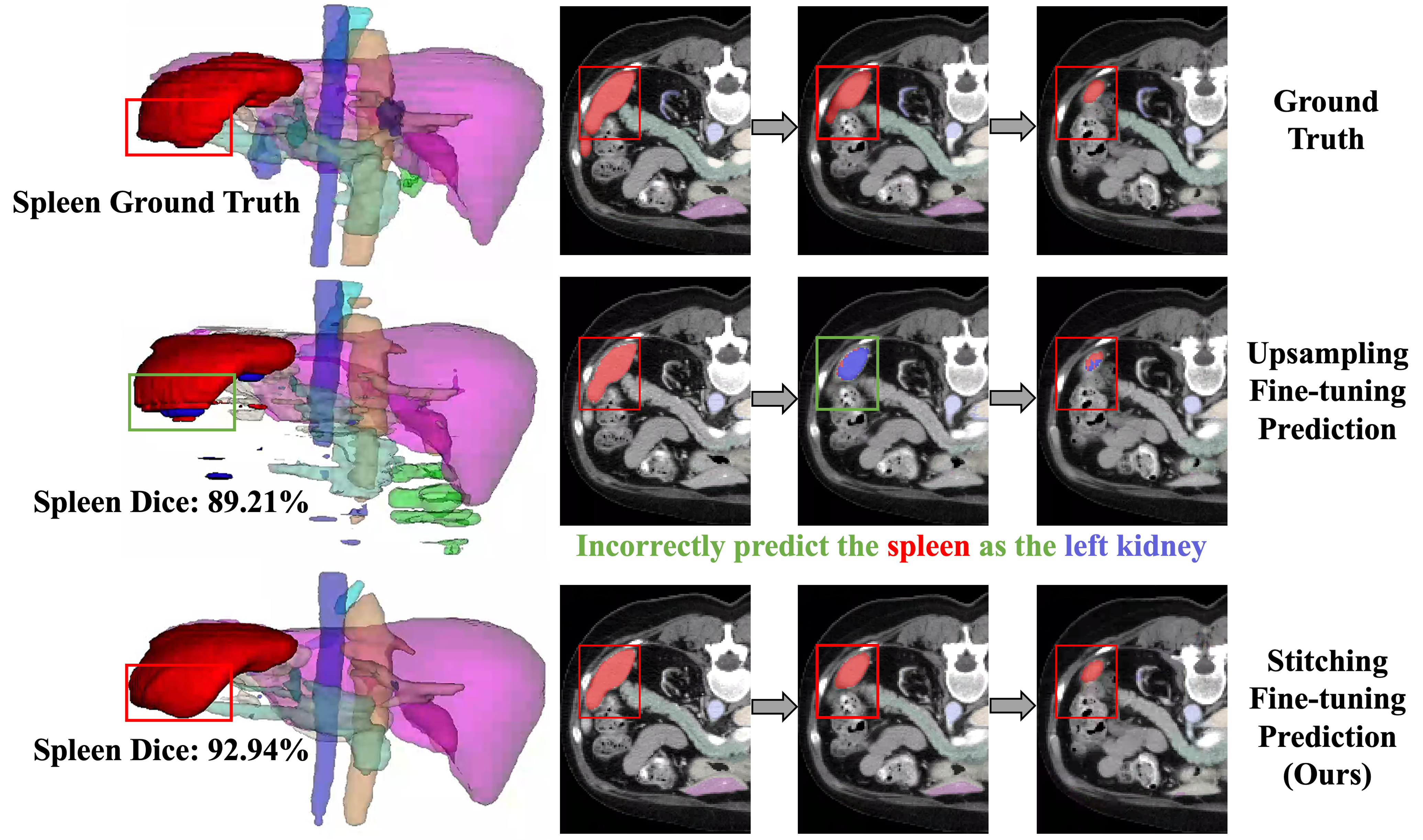}
   \caption{Fine-tuning input strategies comparison. Upsampling fine-tuning directly upsamples each slice.}
   \label{fig: ft_area}
\end{figure}

\begin{figure*}[t]
    \centering
    \includegraphics[width = 0.9\textwidth]{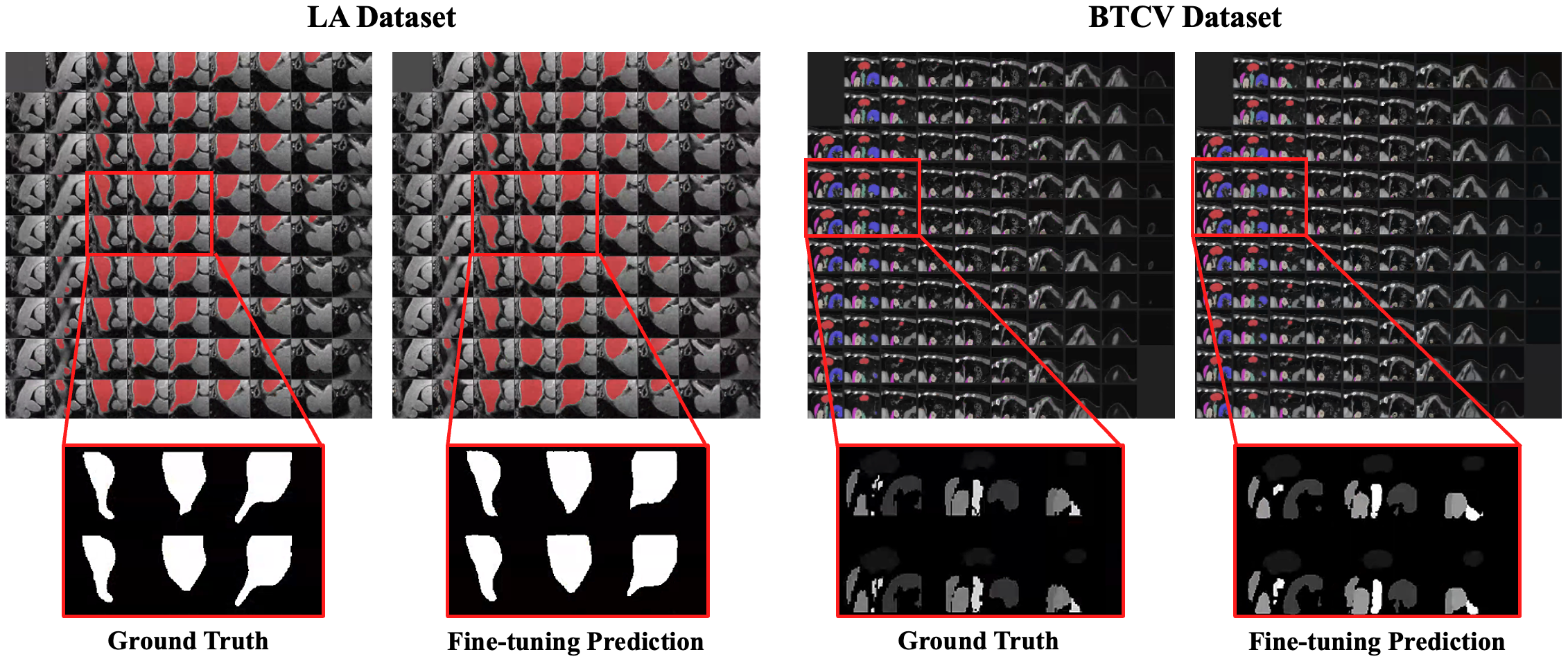}
    \caption{Visualization of fine-tuning module on LA~\cite{xiong2021global} and BTCV~\cite{landman2015miccai} dataset.}
    \label{fig:ft_visual}
\end{figure*}

LoRA offers a practical and effective approach for fine-tuning across domains, and the combination of LoRA with our specific stitching strategies tailored for medical images ensures that our model could handle the substantial differences between natural and medical image data well.

We notice that stitching 2D slices may lead to organs appearing in surrounding and similar regions, guiding SAM to capture these similarities. Our stitching strategy preserves the spatial relationship between slices and could leverage the information of the same target from neighboring slices effectively. For example, as shown in Fig.~\ref{fig: ft_area}, the same semantic class (\eg, spleen) appears in a similar area (upper left) across three adjacent slices. The upsampling fine-tuning method, which predicts slice-by-slice, confuses the entire spleen (red) into the left kidney (blue) in the middle slice. In contrast, our stitching fine-tuning method could correctly identify the spleen in all three slices. By incorporating inter-slice information, our method effectively guides SAM in recognizing the same organ on different slices.
In contrast, our stitching strategy preserves the spatial relationship between slices and could leverage the information of the same target from neighboring slices.

To verify the effectiveness of pseudo-labels on different datasets, we visualize the fine-tuning module prediction of single and multiple target datasets, taking LA~\cite{xiong2021global} and BTCV~\cite{landman2015miccai} datasets as examples, in Fig.~\ref{fig:ft_visual}.
The mask decoder outputs a large-sized 2D mask, which is subsequently restored to a 3D volume as a pseudo-label for the SSL module.

\subsubsection{Re-training SSL Module}
\label{sec:retrain}

As defined above, the training data consists of labeled dataset $L = \left\{ (X^l_i, Y_i) \right\} ^m_{i=1}$ and unlabeled dataset $U = \left\{ X^u_j \right\} ^n_{j=1}$.
The training objective of re-training SSL module $\texttt{F}_\texttt{S}(\cdot)$ can be formulated as:
\begin{equation}
\omega^* = \arg\min_{\omega}(\mathcal{L}_{sup} + \lambda \mathcal{L}_{unsup}), 
\end{equation}
where $\mathcal{L}_{sup}$ and $\mathcal{L}_{unsup}$ are supervised and unsupervised terms, respectively, and $\lambda$ acts as a tradeoff between them. Our pseudo-label guidance $\mathcal{L}_{pl}$ is an unsupervised loss. 
\begin{equation}
\label{eq: pl}
\mathcal{L}_{pl} = \frac{1}{2} \bigl(\mathcal{L}_{Dice} (\mathbf{P}^u_j, \hat{\mathbf{Y}}^u_j) + \mathcal{L}_{ce}(\mathbf{P}^u_j, \hat{\mathbf{Y}}^u_j)\bigr), 
\end{equation}
where $\mathbf{P}^u_j=\texttt{F}_\texttt{S}(\mathbf{X}^u_j)$ is the prediction of SSL module. 

We investigate four 3D medical image semi-supervised methods for our re-training module, including two classical methods (\ie, self-training and mean teacher), and two advanced methods (\ie, ACMT and MagicNet).

\textbf{Self-training}: 
Self-training~\cite{bai2017semi} involves three iteratively steps:  
Firstly, a teacher network is initially trained on $L$. 
Secondly, it makes predictions on $U$ to obtain $\hat{U}$. 
Thirdly, a student network retrained on the union set $L \cup \hat{U}$. 
Among them, steps 2 and 3 are iteratively performed in alternation.

\textbf{Mean Teacher}: 
Mean teacher (MT)~\cite{tarvainen2017mean} consists of a student network and a teacher network, and teacher network weights are updated with the exponential moving average (EMA) of student network weights.  
The unsupervised loss is the consistency regularization between the two networks. 

\textbf{ACMT}: 
ACMT~\cite{xu2023ambiguity} improves MT to the ambiguity-consensus mean teacher model, encouraging consistency between the student’s and the teacher’s predictions at the identified ambiguous regions. 

\begin{figure}[t]
    \centering
    \includegraphics[width = 1.0\linewidth]{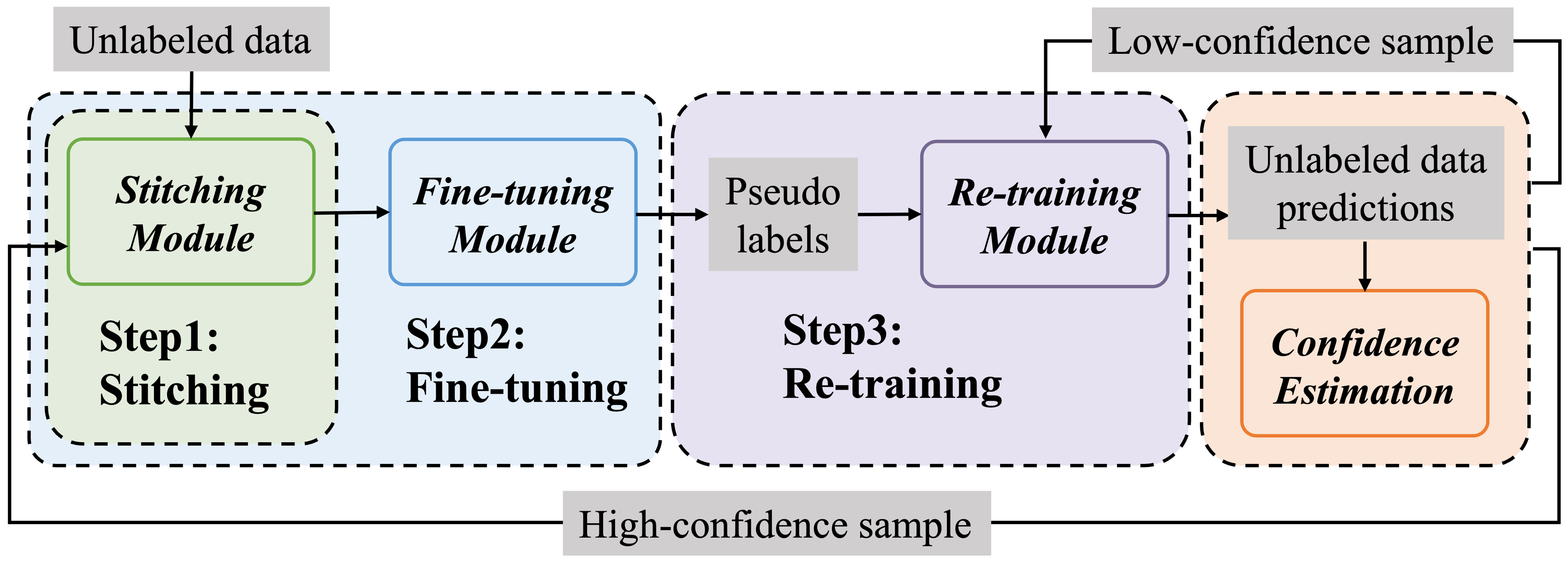}
    \caption{Overview of the proposed SFR$^+$ framework.}
    \label{fig:framework_plus}
\vspace{-5pt}
\end{figure}

\textbf{MagicNet}: 
MagicNet~\cite{chen2023magicnet} introduces a data augmentation strategy based on MT. 
First, a pair of labeled and unlabeled samples are mixed into two shuffled cubes.
Next, small cubes and mixed cubes are fed into the segmentation network, and finally recovery the mixed cubes. 

\subsection{SFR$^+$ Framework}

To further enhance our framework, we developed the extension version of SFR, named SFR$^+$, as illustrated in Fig.~\ref{fig:framework_plus}. In the SFR framework, the fine-tuning module trained with labeled samples provides pseudo-labels of unlabeled samples uni-directionally to the re-training module. To more effectively extract and leverage information of unlabeled data from the two modules, our SFR$^+$ introduces a confidence estimation strategy to distinguish between confident and uncertain samples. This approach enables selective optimization of both the fine-tuning and re-training modules.

\subsubsection{Confidence Estimation}
SFR$^+$ introduces confidence estimation to determine how to handle each unlabeled sample. We calculate the voxel-level average confidence for each unlabeled sample and classify them based on a threshold. For each unlabeled sample $\mathbf{X}^u_j$, the voxel-level average confidence is defined as:
\begin{equation}
\mathbf{C}^u_j = \frac{1}{N_v} \sum_{v} \max_c \, \text{softmax}_c(\texttt{F}_\texttt{S}(\mathbf{X}^u_{j, v})),
\end{equation}
where $\texttt{F}_\texttt{S}(\mathbf{X}^u_{j, v})$ is the model prediction for voxel $v$ and $c$ indexes over the classes. $N_v$ is the number of voxels in $\mathbf{X}^u_j$. 
Samples are classified based on a threshold $\tau$: 
\begin{itemize}
    \item High-confident samples: If $\mathbf{C}^u_j \geq \tau$, the sample is considered confident and to update the fine-tuning module.
    \item Low-confident samples: If $\mathbf{C}^u_j < \tau$, the sample is classified as uncertain and sent to the re-training module.
\end{itemize}

\subsubsection{Selective Training Strategy}
In SFR$^+$, selective learning in the fine-tuning and re-training modules allows for more effective handling of unlabeled samples. 
On one hand, high-confident samples are employed to update the fine-tuning module, ensuring that only reliable information from unlabeled data contributes to further refinement. 
On the other hand, low-confidence samples benefit from the pseudo-labels, which enable further improvements of the re-training module.

By selectively alternating updates between the two modules, SFR$^+$ mitigates the risk of error propagation from inaccurate predictions that may arise due to the uni-directional transfer of pseudo-labels.

\subsection{Summary}
\label{sec:summary}

The training procedure of our SFR framework is summarized in Fig.~\ref{fig: method}.
During fine-tuning, the 3D labeled volumes are first transformed into larger-sized 2D images by the stitching module. Then fine-tuning module is updated by minimizing the supervised fine-tuning loss. During re-training, the fine-tuning module annotates unlabeled samples and provides pseudo-labels to re-training module. The SSL re-training module is trained by incorporating the supervised information from labeled images and the pseudo-label consistency from unlabeled images.

Our SFR and SFR$^+$ framework ensures computational efficiency. The fine-tuning module SFR$_\text{FT}$ with the LoRA strategy has the same parameter size as the foundation model (SAM). The re-training module has a parameter scale to the mainstream segmenters like V-Net~\cite{milletari2016v}. During inference, we discard fine-tuning module and retain only re-training module. The results of new samples are directly predicted using $\texttt{F}_\texttt{S}(\cdot)$.

\section{Experiments}
\label{sec:experiments}

The experiments are conducted on two single target datasets (\ie, LA~\cite{xiong2021global} and BraTS~\cite{menze2014multimodal}) and three multiple target datasets (\ie, BTCV~\cite{landman2015miccai}, MACT~\cite{gibson2018automatic}, and AbdomenCT-1K~\cite{ma2021abdomenct}).
We perform experiments of semi-supervised segmentation with moderate annotations and scarce annotations on each dataset.
Subsequently, we further analyze the pseudo-labels generated by the fine-tuning module, as well as the effectiveness and compatibility of the retraining module, and conduct experiments on different labeled samples. 

%%%%%%%%%%%%%%%%%%%%% subsection %%%%%%%%%%%%%%%%%%%%%%%%
\subsection{Datasets}
\textbf{LA Dataset. }
The LA dataset~\cite{xiong2021global} in the MICCAI 2018 Atrium Segmentation Challenge is for left atrium segmentation in 3D gadolinium-enhanced MR image scans (GE-MRIs). 
It contains 100 scans with an isotropic resolution of $0.625 \times 0.625 \times 0.625$ $\text{mm}^3$, and ground truth masks segmented by expert radiologists.
Fairly, we follow the same data split and pre-processing procedures as the existing work~\cite{yu2019uncertainty, luo2021semi, xu2023ambiguity}. 

\textbf{BraTS Dataset. }
The dataset contains preoperative MRI (with T1, T1Gd, T2 and T2-FLAIR modalities) of 335 glioma patients from the BraTS 2019 challenge~\cite{menze2014multimodal, bakas2017advancing, bakas2018identifying}, where 259 patients with high-grade glioma and 76 with low-grade glioma. 
Following~\cite{xu2023ambiguity}, we only use T2-FLAIR images with the same data split and pre-processing procedures for fair comparison. 

\textbf{BTCV Dataset. }
The BTCV multiorgan dataset~\cite{landman2015miccai} from the MICCAI Multi-Atlas Labeling Beyond Cranial Vault-Workshop Challenge contains 30 subjects with 3779 axial abdominal CT slices. It consists of 13 organ annotations, including 8 organs of Synapse. 
We strictly follow the same data split and pre-processing procedures as the existing work~\cite{chen2023magicnet}, where the volume is divided into $96\times 96\times 96$ patches. 

\textbf{MACT Dataset. }
The MACT dataset~\cite{gibson2018automatic} is a public multi-organ abdominal CT reference standard segmentation dataset,  containing 90 CT volumes with 8 organs annotation. The original data is from the Cancer Image Archive (TCIA) Pancreas-CT dataset and the BTCV dataset. 
We follow the same pre-processing procedure as~\cite{chen2023magicnet}, and we divide 70 cases for training and 20 cases for testing. Following~\cite{chen2023magicnet}, the volume is divided into $96\times 96\times 96$ patches as input volumes. 

\textbf{AbdomenCT-1K Dataset. }
The AbdomenCT-1K dataset~\cite{ma2021abdomenct} is a diverse abdominal CT organ segmentation dataset, with more than 1,000 CT scans from 12 medical centers, including multi-phase, multi-vendor, and multi-disease cases. We follow the same pre-processing procedure as nnU-Net~\cite{isensee2021nnu}. Similar to BTCV and MACT datasets, we divide the image into $96\times 96\times 96$ patches as input volumes.

%%%%%%%%%%%%%%%%%%%%% subsection %%%%%%%%%%%%%%%%%%%%%%%%
\subsection{Experimental Settings}
In this paper, all the experiments are implemented in Pytorch on the NVIDIA GeForce RTX 3090/4090TI GPU. 
For foundation model SAM, we conduct all the experiments based on the “ViT-B" version. We adopt LoRA finetuning and the rank of LoRA is set to 4 for efficiency and performance optimization. 
In our experiments, we follow the current popular setting~\cite{zhang2023customized} in designing mask decoder, which modifies the segmentation head to generate masks of each class in a deterministic manner and aggregate to final segmentation map.

For the semi-supervised network, it is trained by the SGD optimizer with an initial learning rate of 0.01. 
For LA and BraTS datasets, we follow the training strategy of~\cite{yu2019uncertainty, xu2023ambiguity}.
And we employ four measurements to quantitatively evaluate the segmentation performance, including Dice, Jaccard, the average surface distance (ASD), and the 95\% Hausdorff Distance (HD).
For BTCV, MACT, and AbdomenCT-1K datasets, we follow the implementation details of Magicnet~\cite{chen2023magicnet} and Dice as an evaluation metric for multi-organ segmentation.
For fair comparison, we use official reported $\lambda$ of various baselines~\cite{yu2019uncertainty, xu2023ambiguity, chen2023magicnet}. 
In SFR$^+$, we set the $\tau$ as 0.985.

During inference, we discard fine-tuning module and new samples are directly predicted by re-training module. 

\begin{figure}[t]
    \centering
    \includegraphics[width = 0.8\linewidth]{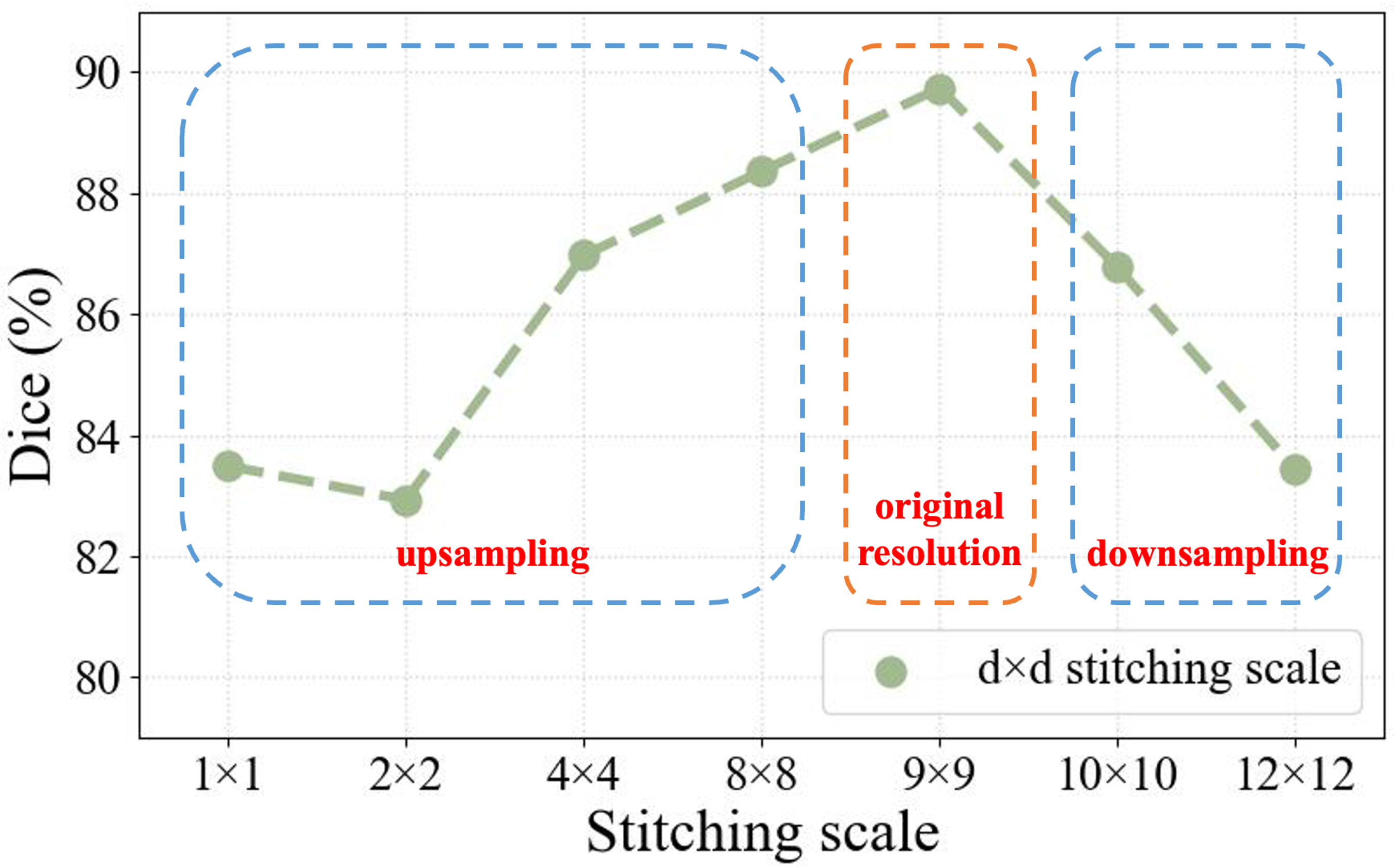}
    \caption{Fine-tuning with different stitching scales on LA dataset with 16 labeled data. When the stitch scale is set to $9 \times 9$, the large-sized 2D image represents the complete 3D volume. }
    \label{fig:scales}
\end{figure}

%%%%%%%%%%%%%%%%%%%%% subsection %%%%%%%%%%%%%%%%%%%%%%%%
\subsection{Stitching Strategy Analysis}
There is a performance trade-off between the resolution of individual slices and the number of slices stitched together into a single 1024 $\times$ 1024 image. Increasing the number of slices provides more context and spatial continuity. However, it also requires resizing the slices to fit into the fixed resolution, potentially resulting in a loss of fine-grained details. Therefore, how to consider both the slice resolution and the stitching scale becomes a critical issue. 

\begin{table}[t]
\caption{Results of LA dataset with moderate annotations. ``L/U" indicates the number of labeled and unlabeled volumes. $\uparrow$ means higher values are better and $\downarrow$ means lower values are preferable. LB and UB are the lower and upper bound, respectively. Metrics are Dice (\%), Jaccard (\%), ASD (voxel), and 95HD (voxel). }
\label{tab:la_ssl}
\centering
\scalebox{0.95}{
\begin{tabular}{c|c|cccc}
\hline
\noalign{\smallskip}
L / U & Method & Dice$\uparrow$ & Jaccard$\uparrow$ & ASD$\downarrow$ & 95HD$\downarrow$ \\
\noalign{\smallskip}
\hline
\noalign{\smallskip}
\multirow{1}{*}{16 / 0} & V-Net\scalebox{0.7}{~\cite{milletari2016v} (LB)} & 86.03 & 76.06 & 3.51 & 14.26 \\ 
\noalign{\smallskip}
\hline
\noalign{\smallskip}
\multirow{15}{*}{16 / 64} & UA-MT \scalebox{0.7}{[MICCAI'19]~\cite{yu2019uncertainty}} & 88.88 & 80.21 & 2.26 & 7.32 \\
& CCT \scalebox{0.7}{[CVPR'20]~\cite{ouali2020semi}} & 88.01 & 80.95 & 2.37 & 8.25 \\
& CPS \scalebox{0.7}{[CVPR'21]~\cite{chen2021semi}} & 87.87 & 78.61 & 2.16 & 12.87 \\
& DTC \scalebox{0.7}{[AAAI'21]~\cite{luo2021semi}} & 89.42 & 80.98 & 2.10 & 7.32 \\
& ICT \scalebox{0.7}{[NN'22]~\cite{verma2022interpolation}} & 89.02 & 80.34 & 1.97 & 10.38 \\
& CPCL \scalebox{0.7}{[JBHI'22]~\cite{xu2022all}} & 88.32 & 81.02 & 2.02 & 8.01 \\
& URPC \scalebox{0.7}{[MedIA'22]~\cite{luo2022semi}} & 88.43 & 81.15 & 2.23 & 8.21 \\
& BCP \scalebox{0.7}{[CVPR'23]~\cite{bai2023bidirectional}} & 90.10 & 82.11 & 2.51 & 7.62 \\
& CauSSL \scalebox{0.7}{[ICCV'23]~\cite{miao2023caussl}} & 89.48 & 81.20 & 1.75 & 7.55 \\
& \cellcolor{brown!20}MT \scalebox{0.7}{[NIPS'17]~\cite{tarvainen2017mean}} & \cellcolor{brown!20}88.12 & \cellcolor{brown!20}79.03 & \cellcolor{brown!20}2.65 & \cellcolor{brown!20}10.92 \\
& \cellcolor{brown!20}\textbf{SFR$_\text{MT}$} \scalebox{0.7}{(Ours)} & \cellcolor{brown!20}90.86 & \cellcolor{brown!20}83.34 & \cellcolor{brown!20}1.45 & \cellcolor{brown!20}6.15 \\
& \cellcolor{brown!20}\textbf{SFR$_\text{MT}^+$} \scalebox{0.7}{(Ours)} & \cellcolor{brown!20}90.91 & \cellcolor{brown!20}83.42 & \cellcolor{brown!20}1.72 & \cellcolor{brown!20}\textbf{5.80} \\
& \cellcolor{teal!20}ACMT \scalebox{0.7}{[MedIA'23]~\cite{xu2023ambiguity}} & \cellcolor{teal!20}90.31 & \cellcolor{teal!20}82.43 & \cellcolor{teal!20}1.76 & \cellcolor{teal!20}6.21 \\
& \cellcolor{teal!20}\textbf{SFR$_\text{ACMT}$} \scalebox{0.7}{(Ours)} & \cellcolor{teal!20}90.95 & \cellcolor{teal!20}83.47 & \cellcolor{teal!20}\textbf{1.43} & \cellcolor{teal!20}6.11 \\
& \cellcolor{teal!20}\textbf{SFR$_\text{ACMT}^+$} \scalebox{0.7}{(Ours)} & \cellcolor{teal!20}\textbf{91.00} & \cellcolor{teal!20}\textbf{83.53} & \cellcolor{teal!20}1.61 & \cellcolor{teal!20}\textbf{6.13} \\
\noalign{\smallskip}
\hline
\noalign{\smallskip}
80 / 0 & V-Net\scalebox{0.7}{~\cite{milletari2016v} (UB)} & 91.14 & 83.82 & 1.52 & 5.75 \\
\noalign{\smallskip}
\hline
\end{tabular}}
\end{table}

In Fig.~\ref{fig:scales}, we explored different stitching scales ($d \times d$ grid) to analyze this trade-off. On the LA dataset~\cite{xiong2021global}, when the stitch scale is set to $9 \times 9$, the large-sized 2D image represents the complete 3D volume, with each slice maintaining its original resolution without any downsampling. The results show that when the stitch scale is small, performance tends to be inferior. With a small scale, it is possible that the representation may not fully capture the volume context, limiting the model's ability to leverage inter-slice information. By increasing the number of slices, the model benefits from more context, resulting in improved segmentation performance, particularly when the stitched image encompasses the entire 3D volume. 
When the resolution of individual slices is maintained, the large-sized 2D image represents the 3D volume without any downsampling, and the best results are achieved.
We supplement this figure by further increasing the stitching scale, that is, reducing the resolution of each slice. However, it becomes counterproductive when further increasing the stitching scale—excessive downsampling of individual slices begins to degrade performance as the resolution becomes too low to capture critical anatomical details.
In our method, each slice retains its resolution throughout the process, ensuring that no upsampling or downsampling occurs, which balances the need for spatial context with the preservation of resolution. The stitching process does not affect the resolution of the individual slices; rather, it combines them into a large-sized 2D image. As a result, the final 2D segmentation predictions are also aligned with the original slice resolution, preserving the fine-grained details essential for high-quality segmentation.

\begin{table}[t]
\caption{Results of BraTS dataset with moderate annotations. }
\label{tab:brats_ssl}
\centering
\scalebox{0.95}{
\begin{tabular}{c|c|cccc}
\hline
\noalign{\smallskip}
L / U & Method & Dice$\uparrow$ & Jaccard$\uparrow$ & ASD$\downarrow$ & 95HD$\downarrow$ \\
\noalign{\smallskip}
\hline
\noalign{\smallskip}
\multirow{1}{*}{50 / 0} & 3D U-Net\scalebox{0.7}{~\cite{cciccek20163d} (LB)} & 80.16 & 71.55 & 3.43 & 22.68 \\
\noalign{\smallskip}
\hline
\noalign{\smallskip}
\multirow{15}{*}{50 / 200} & UA-MT \scalebox{0.7}{[MICCAI'19]~\cite{yu2019uncertainty}} & 83.12 & 73.01 & 2.30 & 9.87  \\
& CCT \scalebox{0.7}{[CVPR'20]~\cite{ouali2020semi}} & 82.53 & 72.36 & 2.21 & 15.87 \\
& CPS \scalebox{0.7}{[CVPR'21]~\cite{chen2021semi}} & 84.01 & 74.02 & 2.18 & 12.16 \\
& DTC \scalebox{0.7}{[AAAI'21]~\cite{luo2021semi}} & 83.43 & 73.56 & 2.34 & 14.77 \\
& ICT \scalebox{0.7}{[NN'22]~\cite{verma2022interpolation}} & 81.76 & 72.01 & 2.82 & 9.66 \\
& CPCL \scalebox{0.7}{[JBHI'22]~\cite{xu2022all}} & 83.48 & 74.08 & 2.08 & 9.53  \\
& URPC \scalebox{0.7}{[MedIA'22]~\cite{luo2022semi}} & 82.93 & 72.57 & 4.19 & 15.93 \\
& BCP \scalebox{0.7}{[CVPR'23]~\cite{bai2023bidirectional}} & 84.17 & 74.37 & 3.24 & 11.69 \\
& CauSSL \scalebox{0.7}{[ICCV'23]~\cite{miao2023caussl}} & 81.09 & 71.01 & 3.76 & 11.90 \\
& \cellcolor{brown!20}MT \scalebox{0.7}{[NIPS'17]~\cite{tarvainen2017mean}} & \cellcolor{brown!20}82.96 & \cellcolor{brown!20}72.95 & \cellcolor{brown!20}2.32 & \cellcolor{brown!20}9.85 \\
& \cellcolor{brown!20}\textbf{SFR$_\text{MT}$} \scalebox{0.7}{(Ours)} & \cellcolor{brown!20}85.19 & \cellcolor{brown!20}75.77 & \cellcolor{brown!20}2.92 & \cellcolor{brown!20}11.04 \\
& \cellcolor{brown!20}\textbf{SFR$_\text{MT}^+$} \scalebox{0.7}{(Ours)} & \cellcolor{brown!20}86.08 & \cellcolor{brown!20}76.79 & \cellcolor{brown!20}1.93 & \cellcolor{brown!20}8.50 \\
& \cellcolor{teal!20}ACMT \scalebox{0.7}{[MedIA'23]~\cite{xu2023ambiguity}} & \cellcolor{teal!20}84.63 & \cellcolor{teal!20}74.39 & \cellcolor{teal!20}2.11 & \cellcolor{teal!20}9.50 \\
& \cellcolor{teal!20}\textbf{SFR$_\text{ACMT}$} \scalebox{0.7}{(Ours)} & \cellcolor{teal!20}85.81 & \cellcolor{teal!20}76.66 & \cellcolor{teal!20}\textbf{1.79} & \cellcolor{teal!20}\textbf{7.75} \\ 
& \cellcolor{teal!20}\textbf{SFR$_\text{ACMT}^+$} \scalebox{0.7}{(Ours)} & \cellcolor{teal!20}\textbf{86.09} & \cellcolor{teal!20}\textbf{76.80} & \cellcolor{teal!20}2.47 & \cellcolor{teal!20}8.51 \\ 
\noalign{\smallskip}
\hline
\noalign{\smallskip}
250 / 0 & 3D U-Net\scalebox{0.7}{~\cite{cciccek20163d} (UB)} & 85.93 & 76.81 & 1.93 & 9.85 \\ 
\noalign{\smallskip}
\hline
\end{tabular}}
\end{table}

\begin{table}[t]
\caption{Results of AbdomenCT-1K dataset with moderate annotations and the evaluation metric is the Dice score (\%). * means all the pixels are predicted as background or another region.}
\label{tab:abdomen_ssl}
\centering
\scalebox{0.83}{
\textcolor{black}{
\begin{tabular}{c|c|c|cccc}
\hline
\noalign{\smallskip}
L / U & Method & AVG $\uparrow$ & Liver & Kidney & Spleen & Pancreas \\ 
\noalign{\smallskip}
\hline
\noalign{\smallskip}
\multirow{1}{*}{180 / 0} & V-Net\scalebox{0.7}{~\cite{milletari2016v} (LB)} & 67.17 & 96.03 & 93.64 & 78.99 & 0* \\
\noalign{\smallskip}
\hline
\noalign{\smallskip}
\multirow{6}{*}{180 / 720} 
& \cellcolor{brown!20}MT \scalebox{0.7}{[NIPS'17]~\cite{tarvainen2017mean}} & \cellcolor{brown!20}75.16 & \cellcolor{brown!20}93.78 & \cellcolor{brown!20}91.54 & \cellcolor{brown!20}76.55 & \cellcolor{brown!20}38.79 \\
& \cellcolor{brown!20}\textbf{SFR$_\text{MT}$ \scalebox{0.7}{(Ours)}} & \cellcolor{brown!20}88.73 & \cellcolor{brown!20}96.10 & \cellcolor{brown!20}94.55 & \cellcolor{brown!20}93.63 & \cellcolor{brown!20}70.65 \\
& \cellcolor{brown!20}\textbf{SFR$_\text{MT}^+$ \scalebox{0.7}{(Ours)}} & \cellcolor{brown!20}90.00 & \cellcolor{brown!20}96.11 & \cellcolor{brown!20}94.50 & \cellcolor{brown!20}95.19 & \cellcolor{brown!20}74.19 \\
& \cellcolor{teal!20}MagicNet \scalebox{0.7}{[CVPR'23]~\cite{chen2023magicnet}} & \cellcolor{teal!20}90.54 & \cellcolor{teal!20}96.38 & \cellcolor{teal!20}94.54 & \cellcolor{teal!20}95.17 & \cellcolor{teal!20}76.08 \\
& \cellcolor{teal!20}\textbf{SFR$_\text{MagicNet}$ \scalebox{0.7}{(Ours)}} & \cellcolor{teal!20}91.41 & \cellcolor{teal!20}\textbf{96.70} & \cellcolor{teal!20}95.01 & \cellcolor{teal!20}96.02 & \cellcolor{teal!20}77.90 \\ 
& \cellcolor{teal!20}\textbf{SFR$_\text{MagicNet}^+$ \scalebox{0.7}{(Ours)}} & \cellcolor{teal!20}\textbf{91.70} & \cellcolor{teal!20}96.66 & \cellcolor{teal!20}\textbf{95.20} & \cellcolor{teal!20}\textbf{96.06} & \cellcolor{teal!20}\textbf{78.88} \\
\noalign{\smallskip}
\hline
\noalign{\smallskip}
900 / 0 & V-Net\scalebox{0.7}{~\cite{milletari2016v} (UB)} & 88.26 & 95.31 & 94.49 & 92.30 & 70.94 \\
\noalign{\smallskip}
\hline
\end{tabular}}}
\end{table}

\begin{table*}[t]
\caption{Results of BTCV dataset with moderate annotations and the evaluation metric is the Dice score (\%). Note: Spl: spleen, R.Kid: right kidney, L.Kid: left kidney, Gall: gallbladder, Eso: esophagus, Liv: liver, Sto: stomach, Aor: aorta, IVC: inferior vena cava, Veins: portal and splenic veins, Pan: pancreas, LG/RG: left/right adrenal glands. }
\label{tab:btcv_ssl}
\centering
\scalebox{0.92}{
\begin{tabular}{c|c|c|ccccccccccccc}
\hline
\noalign{\smallskip}
L / U & Method & AVG $\uparrow$ & Spl & R.Kid & L.Kid & Gall & Eso & Liv & Sto & Aor & IVC & Veins & Pan & RG & LG \\ 
\noalign{\smallskip}
\hline
\noalign{\smallskip}
\multirow{1}{*}{7 / 0} & V-Net\scalebox{0.7}{~\cite{milletari2016v} (LB)} & 67.17 & 84.98 & 82.72 & 82.07 & 36.64 & 63.48 & 93.54 & 57.49 & 89.74 & 78.63 & 60.42 & 49.39 & 55.60 & 38.49 \\
\noalign{\smallskip}
\hline
\noalign{\smallskip}
\multirow{11}{*}{7 / 11} & UA-MT \scalebox{0.7}{[MICCAI'19]~\cite{yu2019uncertainty}} & 67.75 & 88.74 & 75.88 & 78.91 & 54.25 & 58.55 & 93.46 & 58.90 & 89.23 & 76.15 & 62.30 & 47.91 & 51.53 & 44.92 \\
& CPS \scalebox{0.7}{[CVPR'21]~\cite{chen2021semi}} & 65.81 & 87.56 & 72.99 & 77.59 & 53.31 & 54.08 & 92.41 & 54.58 & 87.75 & 74.32 & 58.68 & 48.02 & 50.39 & 43.86 \\
& ICT \scalebox{0.7}{[NN'22]~\cite{verma2022interpolation}} & 73.69 & 90.31 & 84.41 & 86.96 & 49.22 & 65.65 & 94.29 & 65.95 & 90.23 & 81.44 & 69.56 & 66.61 & 57.35 & 56.01 \\
& SS-Net \scalebox{0.7}{[MICCAI'22]~\cite{wu2022exploring}} & 58.26 & 84.74 & 76.37 & 74.19 & 43.42 & 57.05 & 92.90 & 14.37 & 83.14 & 69.77 & 52.45 & 27.08 & 54.29 & 27.66 \\
& SLC-Net \scalebox{0.7}{[MICCAI'22]~\cite{liu2022semi}} & 70.40 & 90.05 & 84.00 & 86.43 & 56.16 & 58.91 & 94.68 & 70.72 & 89.93 & 79.45 & 60.59 & 54.22 & 51.03 & 39.08 \\
& \cellcolor{brown!20}MT \scalebox{0.7}{[NIPS'17]~\cite{tarvainen2017mean}} & \cellcolor{brown!20}65.68 & \cellcolor{brown!20}85.70 & \cellcolor{brown!20}78.93 & \cellcolor{brown!20}79.08 & \cellcolor{brown!20}42.80 & \cellcolor{brown!20}61.09 & \cellcolor{brown!20}93.45 & \cellcolor{brown!20}57.57 & \cellcolor{brown!20}89.70 & \cellcolor{brown!20}80.30 & \cellcolor{brown!20}63.95 & \cellcolor{brown!20}41.14 & \cellcolor{brown!20}50.46 & \cellcolor{brown!20}29.69 \\
& \cellcolor{brown!20}\textbf{SFR$_\text{MT}$} \scalebox{0.7}{(Ours)} & \cellcolor{brown!20}70.09 \scalebox{0.8}{({\color{red}$\uparrow$4.41})} & \cellcolor{brown!20}87.92 & \cellcolor{brown!20}82.86 & \cellcolor{brown!20}81.94 & \cellcolor{brown!20}53.02 & \cellcolor{brown!20}60.50 & \cellcolor{brown!20}95.06 & \cellcolor{brown!20}72.82 & \cellcolor{brown!20}89.71 & \cellcolor{brown!20}81.23 & \cellcolor{brown!20}67.46 & \cellcolor{brown!20}59.29 & \cellcolor{brown!20}41.77 & \cellcolor{brown!20}37.58 \\
& \cellcolor{brown!20}\textbf{SFR$_\text{MT}^+$ \scalebox{0.7}{(Ours)}} & \cellcolor{brown!20}71.81 \scalebox{0.8}{({\color{red}$\uparrow$6.13})} & \cellcolor{brown!20}88.78 & \cellcolor{brown!20}84.08 & \cellcolor{brown!20}86.48 & \cellcolor{brown!20}55.04 & \cellcolor{brown!20}64.61 & \cellcolor{brown!20}\textbf{95.08} & \cellcolor{brown!20}74.60 & \cellcolor{brown!20}89.89 & \cellcolor{brown!20}81.62 & \cellcolor{brown!20}63.38 & \cellcolor{brown!20}63.47 & \cellcolor{brown!20}41.99 & \cellcolor{brown!20}44.57 \\
& \cellcolor{teal!20}MagicNet \scalebox{0.7}{[CVPR'23]~\cite{chen2023magicnet}} & \cellcolor{teal!20}76.74 & \cellcolor{teal!20}91.61 & \cellcolor{teal!20}85.02 & \cellcolor{teal!20}\textbf{88.13} & \cellcolor{teal!20}58.16 & \cellcolor{teal!20}66.72 & \cellcolor{teal!20}94.07 & \cellcolor{teal!20}74.46 & \cellcolor{teal!20}90.77 & \cellcolor{teal!20}84.31 & \cellcolor{teal!20}71.56 & \cellcolor{teal!20}68.90 & \cellcolor{teal!20}63.48 & \cellcolor{teal!20}60.47 \\
& \cellcolor{teal!20}\textbf{SFR$_\text{MagicNet}$} \scalebox{0.7}{(Ours)} & \cellcolor{teal!20}77.06 \scalebox{0.8}{({\color{red}$\uparrow$0.32})} & \cellcolor{teal!20}\textbf{92.09} & \cellcolor{teal!20}85.36 & \cellcolor{teal!20}83.70 & \cellcolor{teal!20}\textbf{63.38} & \cellcolor{teal!20}69.97 & \cellcolor{teal!20}94.15 & \cellcolor{teal!20}74.69 & \cellcolor{teal!20}91.14 & \cellcolor{teal!20}\textbf{84.45} & \cellcolor{teal!20}70.68 & \cellcolor{teal!20}67.22 & \cellcolor{teal!20}\textbf{64.39} & \cellcolor{teal!20}\textbf{60.53} \\ 
& \cellcolor{teal!20}\textbf{SFR$_\text{MagicNet}^+$ \scalebox{0.7}{(Ours)}} & \cellcolor{teal!20}\textbf{77.07} \scalebox{0.8}{({\color{red}$\uparrow$0.33})} & \cellcolor{teal!20}90.43 & \cellcolor{teal!20}\textbf{86.03} & \cellcolor{teal!20}86.15 & \cellcolor{teal!20}60.32 & \cellcolor{teal!20}\textbf{70.72} & \cellcolor{teal!20}94.61 & \cellcolor{teal!20}\textbf{74.83} & \cellcolor{teal!20}\textbf{91.15} & \cellcolor{teal!20}84.21 & \cellcolor{teal!20}\textbf{71.57} & \cellcolor{teal!20}\textbf{70.39} & \cellcolor{teal!20}61.55 & \cellcolor{teal!20}59.90 \\
\noalign{\smallskip}
\hline
\noalign{\smallskip}
18 / 0 & V-Net\scalebox{0.7}{~\cite{milletari2016v} (UB)} & 76.28 & 84.00 & 84.82 & 86.38 & 67.42 & 65.02 & 94.83 & 73.75 & 90.27 & 84.19 & 69.85 & 63.54 & 62.60 & 65.02 \\ 
\noalign{\smallskip}
\hline
\end{tabular}}
\end{table*}

\begin{table*}[t]
\caption{Results of MACT dataset with moderate annotations and the evaluation metric is the Dice score (\%). L.Kidey: left kidney. * means all the pixels are predicted as background or another region.}
\label{tab:mact_ssl}
\centering
\scalebox{0.92}{
\begin{tabular}{c|c|c|cccccccc}
\hline
\noalign{\smallskip}
L / U & Method & AVG $\uparrow$ & Spleen & L.Kedney & Gallbladder & Esophagus & Liver & Stomach & Pancreas & Doudenum \\ 
\noalign{\smallskip}
\hline
\noalign{\smallskip}
\multirow{1}{*}{14 / 0} & V-Net\scalebox{0.7}{~\cite{milletari2016v} (LB)} & 69.05 & 93.94 & 94.36 & 60.43 & 0* & 95.57 & 78.04 & 72.19 & 57.83 \\
\noalign{\smallskip}
\hline
\noalign{\smallskip}
\multirow{10}{*}{14 / 56} & UA-MT \scalebox{0.7}{[MICCAI'19]~\cite{yu2019uncertainty}} & 78.33 & 93.76 & 92.07 & 75.01 & 65.53 & 95.42 & 77.91 & 72.40 & 54.57 \\
& CPS \scalebox{0.7}{[CVPR'21]~\cite{chen2021semi}} & 65.17 & 93.84 & 79.80 & 64.62 & 0* & 93.66 & 81.49 & 62.25 & 45.70 \\
& ICT \scalebox{0.7}{[NN'22]~\cite{verma2022interpolation}} & 77.52 & 93.12 & 92.05 & 71.51 & 67.54 & 94.38 & 77.40 & 68.03 & 56.16 \\
& SS-Net \scalebox{0.7}{[MICCAI'22]~\cite{wu2022exploring}} & 69.69 & 93.15 & 92.89 & 71.75 & 0* & 93.08 & 73.99 & 73.11 & 59.56 \\
& \cellcolor{brown!20}MT \scalebox{0.7}{[NIPS'17]~\cite{tarvainen2017mean}} & \cellcolor{brown!20}77.10 & \cellcolor{brown!20}93.52 & \cellcolor{brown!20}93.06 & \cellcolor{brown!20}75.19 & \cellcolor{brown!20}67.61 & \cellcolor{brown!20}94.44 & \cellcolor{brown!20}67.99 & \cellcolor{brown!20}73.39 & \cellcolor{brown!20}51.56 \\
& \cellcolor{brown!20}\textbf{SFR$_\text{MT}$} \scalebox{0.7}{(Ours)} & \cellcolor{brown!20}82.50 \scalebox{0.8}{({\color{red}$\uparrow$5.40})} & \cellcolor{brown!20}\textbf{95.77} & \cellcolor{brown!20}94.42 & \cellcolor{brown!20}\textbf{83.61} & \cellcolor{brown!20}68.92 & \cellcolor{brown!20}\textbf{96.05} & \cellcolor{brown!20}82.81 & \cellcolor{brown!20}76.30 & \cellcolor{brown!20}62.10 \\
& \cellcolor{brown!20}\textbf{SFR$_\text{MT}^+$ \scalebox{0.7}{(Ours)}} & \cellcolor{brown!20}83.05 \scalebox{0.8}{({\color{red}$\uparrow$5.59})} & \cellcolor{brown!20}95.35 & \cellcolor{brown!20}92.84 & \cellcolor{brown!20}80.67 & \cellcolor{brown!20}69.93 & \cellcolor{brown!20}95.44 & \cellcolor{brown!20}\textbf{87.64} & \cellcolor{brown!20}\textbf{77.32} & \cellcolor{brown!20}65.18 \\
& \cellcolor{teal!20}MagicNet \scalebox{0.7}{[CVPR'23]~\cite{chen2023magicnet}} & \cellcolor{teal!20}81.04 & \cellcolor{teal!20}94.19 & \cellcolor{teal!20}94.38 & \cellcolor{teal!20}76.07 & \cellcolor{teal!20}74.08 & \cellcolor{teal!20}95.04 & \cellcolor{teal!20}78.55 & \cellcolor{teal!20}73.36 & \cellcolor{teal!20}62.61 \\
& \cellcolor{teal!20}\textbf{SFR$_\text{MagicNet}$} \scalebox{0.7}{(Ours)} & \cellcolor{teal!20}82.87 \scalebox{0.8}{({\color{red}$\uparrow$1.83})} & \cellcolor{teal!20}95.54 & \cellcolor{teal!20}\textbf{94.45} & \cellcolor{teal!20}82.67 & \cellcolor{teal!20}72.89 & \cellcolor{teal!20}95.93 & \cellcolor{teal!20}81.49 & \cellcolor{teal!20}75.90 & \cellcolor{teal!20}64.06 \\ 
& \cellcolor{teal!20}\textbf{SFR$_\text{MagicNet}^+$ \scalebox{0.7}{(Ours)}} & \cellcolor{teal!20}\textbf{83.47} \scalebox{0.8}{({\color{red}$\uparrow$2.43})} & \cellcolor{teal!20}94.65 & \cellcolor{teal!20}94.39 & \cellcolor{teal!20}81.63 & \cellcolor{teal!20}\textbf{74.46} & \cellcolor{teal!20}95.15 & \cellcolor{teal!20}84.41 & \cellcolor{teal!20}74.85 & \cellcolor{teal!20}\textbf{68.21} \\
\noalign{\smallskip}
\hline
\noalign{\smallskip}
70 / 0 & V-Net\scalebox{0.7}{~\cite{milletari2016v} (UB)} & 85.85 & 96.07 & 95.05 & 84.09 & 72.28 & 96.31 & 89.45 & 82.05 & 71.47 \\
\noalign{\smallskip}
\hline
\end{tabular}}
\end{table*}

In this work, our stitching module reorganizes a volume (3D raw image or 3D patch) into a $1024 \times 1024$ image, enabling a complete representation of the volume within a single image. For the size of slices, different slices might vary a lot in different datasets. For large-scale medical datasets, such as multi-organ abdominal scans, the full 3D volume often has a very large voxel size (\eg, $512 \times 512 \times 512$). In 3D medical image segmentation approaches like 3D-UNet~\cite{cciccek20163d}, V-Net~\cite{milletari2016v}, and nnU-Net~\cite{isensee2021nnu}, it is common practice to divide the volume into smaller 3D patches (\eg, $96 \times 96 \times 96$) for training and inference. In our approach, we follow a similar strategy during both the fine-tuning and retraining phases, using patches of consistent size. Before fine-tuning SAM, we stitch the 3D patches into a $1024 \times 1024$ image, ensuring that the patch structure is preserved throughout the process. 

Specifically, we divide the full 3D volume into consistent 3D patches, each maintaining an original resolution to avoid any loss of fine anatomical details. Each patch is then systematically stitched into a $1024 \times 1024$ image to capture full spatial context. For the LA segmentation~\cite{xiong2021global}, we adopted the same strategy as works~\cite{yu2019uncertainty, luo2021semi, xu2023ambiguity}, where the patch size is set to $H \times W \times D$ where $W=H=112$ and $D=80$. On the BraTS dataset~\cite{menze2014multimodal}, we followed the approach used in ACMT~\cite{xu2023ambiguity}, using patch dimensions of $W=H=D=96$. For the BTCV~\cite{landman2015miccai}, MACT~\cite{gibson2018automatic} and AbdomenCT-1K~\cite{ma2021abdomenct} datasets, following MagicNet~\cite{chen2023magicnet}, the patch size is set to $W=H=D=96$. After fine-tuning, the predictions are restored to their 3D form through an inverse stitching transform, ensuring that the resolution and spatial integrity of the original 3D image are preserved. By carefully adjusting the patch size and stitching scale according to each dataset’s needs, our approach ensures robust performance across both single-target and multi-target segmentation tasks.

\begin{table}[t]
\caption{Results of LA dataset with $96^3$ patch size.}
\label{tab:la_96}
\centering
\scalebox{0.9}{
\begin{tabular}{c|c|cccc}
\hline
\noalign{\smallskip}
L / U & Method & Dice$\uparrow$ & Jaccard$\uparrow$ & ASD$\downarrow$ & 95HD$\downarrow$ \\
\noalign{\smallskip}
\hline
\noalign{\smallskip}
\multirow{4}{*}{16 / 64} 
& \cellcolor{brown!20}MT \scalebox{0.7}{[NIPS'17]~\cite{tarvainen2017mean}} & \cellcolor{brown!20}87.47 & \cellcolor{brown!20}78.03 & \cellcolor{brown!20}3.92 & \cellcolor{brown!20}14.13 \\
& \cellcolor{brown!20}\textbf{SFR$_\text{MT}$ \scalebox{0.7}{(Ours)}} & \cellcolor{brown!20}89.09 \scalebox{0.8}{({\color{red}$\uparrow$1.62})} & \cellcolor{brown!20}80.73 & \cellcolor{brown!20}2.58 & \cellcolor{brown!20}10.33 \\
& \cellcolor{teal!20}ACMT \scalebox{0.7}{[MedIA'23]~\cite{xu2023ambiguity}} & \cellcolor{teal!20}88.50 & \cellcolor{teal!20}79.56 & \cellcolor{teal!20}3.58 & \cellcolor{teal!20}14.71 \\
& \cellcolor{teal!20}\textbf{SFR$_\text{ACMT}$ \scalebox{0.7}{(Ours)}} & \cellcolor{teal!20}89.49 \scalebox{0.8}{({\color{red}$\uparrow$0.99})} & \cellcolor{teal!20}81.09 & \cellcolor{teal!20}1.94 & \cellcolor{teal!20}7.88 \\
\noalign{\smallskip}
\hline
\noalign{\smallskip}
\multirow{4}{*}{1 / 79} 
& \cellcolor{brown!20}MT \scalebox{0.7}{[NIPS'17]~\cite{tarvainen2017mean}} & \cellcolor{brown!20}44.89 & \cellcolor{brown!20}29.85 & \cellcolor{brown!20}18.86 & \cellcolor{brown!20}44.78 \\
& \cellcolor{brown!20}\textbf{SFR$_\text{MT}$ \scalebox{0.7}{(Ours)}} & \cellcolor{brown!20}65.65 \scalebox{0.8}{({\color{red}$\uparrow$20.76})} & \cellcolor{brown!20}53.64 & \cellcolor{brown!20}9.82 & \cellcolor{brown!20}29.43 \\
& \cellcolor{teal!20}ACMT \scalebox{0.7}{[MedIA'23]~\cite{xu2023ambiguity}} & \cellcolor{teal!20}58.40 & \cellcolor{teal!20}42.61 & \cellcolor{teal!20}20.55 & \cellcolor{teal!20}50.54 \\
& \cellcolor{teal!20}\textbf{SFR$_\text{ACMT}$ \scalebox{0.7}{(Ours)}} & \cellcolor{teal!20}68.24 \scalebox{0.8}{({\color{red}$\uparrow$9.84})} & \cellcolor{teal!20}57.40 & \cellcolor{teal!20}12.38 & \cellcolor{teal!20}30.76 \\
\noalign{\smallskip}
\hline
\end{tabular}}
\end{table}

\begin{figure*}[t]
    \centering
    \includegraphics[width = 0.7\linewidth]{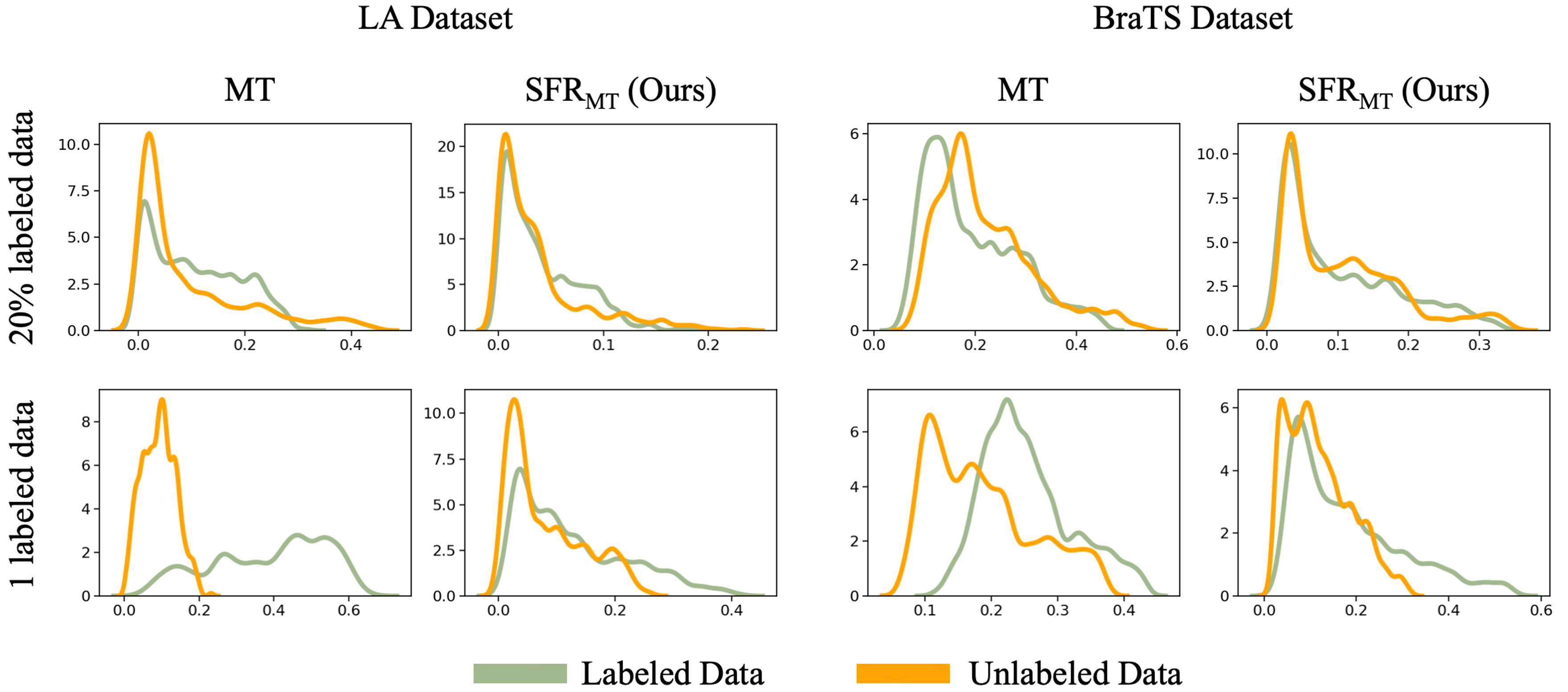}
    \caption{Kernel dense estimations of MT and SFR$_\text{MT}$ on LA and BraTS datasets with Moderate Annotation (20\% labeled data) and Scarce Annotation (1 labeled data).}
    \label{fig:kde_la_brats}
\vspace{-5pt}
\end{figure*}

In addition, we investigate a simplified strategy to ensure uniformity across different datasets and further validate the robustness of our framework. In this alternative approach, we processed all 3D medical images, regardless of their original size, by first cropping into $96 \times 96 \times 96$ patches. These patches are then directly stitched together to form a $10 \times 10$ 2D image. Typically, the resolution of slices (both in height and width) in 3D medical images exceeds 96, though the number of slices (depth) may vary. For datasets where the number of slices is fewer than 96, we employ interpolation to achieve the desired depth. This approach provides a consistent standard for stitching, ensuring that the slices are uniformly processed before being fed into the model, without manual intervention. We have conducted experiments using this approach on the LA dataset with both moderate and scarce annotations, and the results are shown in Table~\ref{tab:la_96}. Notably, our method achieves over 89\% Dice score on moderate annotations and outperforms SSL methods by a large margin on scarce annotations. This demonstrates that even with this simplified and fixed patch-size mechanism, the performance remains effective.

\begin{table}[t]
\caption{Results of LA dataset with scarce annotations. }
\label{tab:la_bsl}
\centering
\scalebox{0.92}{
\begin{tabular}{c|c|cccc}
\hline
\noalign{\smallskip}
L / U & Method & Dice$\uparrow$ & Jaccard$\uparrow$ & ASD$\downarrow$ & 95HD$\downarrow$ \\
\noalign{\smallskip}
\hline
\noalign{\smallskip}
\multirow{1}{*}{1 / 0} & V-Net\scalebox{0.7}{~\cite{milletari2016v} (LB)} & 17.99 & 12.93 & 19.66 & 44.58 \\
\noalign{\smallskip}
\hline
\noalign{\smallskip}
\multirow{6}{*}{1 / 79} 
& \cellcolor{brown!20}MT \scalebox{0.7}{[NIPS'17]~\cite{tarvainen2017mean}} & \cellcolor{brown!20}29.68 & \cellcolor{brown!20}18.19 & \cellcolor{brown!20}18.63 & \cellcolor{brown!20}42.58 \\
& \cellcolor{brown!20}\textbf{SFR$_\text{MT}$} \scalebox{0.7}{(Ours)} & \cellcolor{brown!20}74.40 \scalebox{0.8}{({\color{red}$\uparrow$44.72})} & \cellcolor{brown!20}61.47 & \cellcolor{brown!20}6.25 & \cellcolor{brown!20}25.90 \\
& \cellcolor{brown!20}\textbf{SFR$_\text{MT}^+$} \scalebox{0.7}{(Ours)} & \cellcolor{brown!20}81.03 \scalebox{0.8}{({\color{red}$\uparrow$51.35})} & \cellcolor{brown!20}69.44 & \cellcolor{brown!20}5.12 & \cellcolor{brown!20}19.77 \\
& \cellcolor{teal!20}ACMT \scalebox{0.7}{[MedIA'23]~\cite{xu2023ambiguity}} & \cellcolor{teal!20}72.64 & \cellcolor{teal!20}58.33 & \cellcolor{teal!20}10.42 & \cellcolor{teal!20}33.22 \\
& \cellcolor{teal!20}\textbf{SFR$_\text{ACMT}$} \scalebox{0.7}{(Ours)} & \cellcolor{teal!20}76.76 \scalebox{0.8}{({\color{red}$\uparrow$4.12})} & \cellcolor{teal!20}\textbf{64.65} & \cellcolor{teal!20}8.54 & \cellcolor{teal!20}28.07 \\ 
& \cellcolor{teal!20}\textbf{SFR$_\text{ACMT}^+$} \scalebox{0.7}{(Ours)} & \cellcolor{teal!20}\textbf{83.72} \scalebox{0.8}{({\color{red}$\uparrow$11.08})} & \cellcolor{teal!20}\textbf{72.80} & \cellcolor{teal!20}\textbf{4.54} & \cellcolor{teal!20}\textbf{18.60} \\ 
\noalign{\smallskip}
\hline
\end{tabular}}
\end{table}

\begin{table}[t]
\caption{Results of BraTS dataset with scarce annotations.}
\label{tab:brats_bsl}
\centering
\scalebox{0.92}{
\begin{tabular}{c|c|cccc}
\hline
\noalign{\smallskip}
L / U & Method & Dice$\uparrow$ & Jaccard$\uparrow$ & ASD$\downarrow$ & 95HD$\downarrow$ \\
\noalign{\smallskip}
\hline
\noalign{\smallskip}
\multirow{1}{*}{1 / 0} & 3D U-Net\scalebox{0.7}{~\cite{cciccek20163d} (LB)} & 73.74 & 61.44 & 13.81 & 37.07 \\
\noalign{\smallskip}
\hline
\noalign{\smallskip}
\multirow{6}{*}{1 / 249}
& \cellcolor{brown!20}MT \scalebox{0.7}{[NIPS'17]~\cite{tarvainen2017mean}} & \cellcolor{brown!20}63.52 & \cellcolor{brown!20}50.59 & \cellcolor{brown!20}20.59 & \cellcolor{brown!20}47.54 \\
& \cellcolor{brown!20}\textbf{SFR$_\text{MT}$} \scalebox{0.7}{(Ours)} & \cellcolor{brown!20}78.58 \scalebox{0.8}{({\color{red}$\uparrow$15.06})} & \cellcolor{brown!20}66.56 & \cellcolor{brown!20}7.16 & \cellcolor{brown!20}23.43 \\
& \cellcolor{brown!20}\textbf{SFR$_\text{MT}^+$} \scalebox{0.7}{(Ours)} & \cellcolor{brown!20}79.04 \scalebox{0.8}{({\color{red}$\uparrow$15.52})} & \cellcolor{brown!20}67.55 & \cellcolor{brown!20}\textbf{6.06} & \cellcolor{brown!20}\textbf{21.11} \\
& \cellcolor{teal!20}ACMT \scalebox{0.7}{[MedIA'23]~\cite{xu2023ambiguity}} & \cellcolor{teal!20}63.32 & \cellcolor{teal!20}51.41 & \cellcolor{teal!20}9.28 & \cellcolor{teal!20}31.71 \\
& \cellcolor{teal!20}\textbf{SFR$_\text{ACMT}$} \scalebox{0.7}{(Ours)} & \cellcolor{teal!20}78.47 \scalebox{0.8}{({\color{red}$\uparrow$15.15})} & \cellcolor{teal!20}66.33 & \cellcolor{teal!20}8.17 & \cellcolor{teal!20}26.64 \\
& \cellcolor{teal!20}\textbf{SFR$_\text{ACMT}^+$} \scalebox{0.7}{(Ours)} & \cellcolor{teal!20}\textbf{79.24} \scalebox{0.8}{({\color{red}$\uparrow$15.92})} & \cellcolor{teal!20}\textbf{67.56} & \cellcolor{teal!20}6.72 & \cellcolor{teal!20}22.88 \\
\noalign{\smallskip}
\hline
\end{tabular}}
\end{table}

\begin{table}[t]
\caption{Results of AbdomenCT-1K dataset with scarce annotations and the evaluation metric is the Dice score (\%). * means all the pixels are predicted as background or another region.}
\label{tab:abdomen_bsl}
\centering
\scalebox{0.85}{
\textcolor{black}{
\begin{tabular}{c|c|c|cccc}
\hline
\noalign{\smallskip}
L / U & Method & AVG $\uparrow$ & Liver & Kidney & Spleen & Pancreas \\ 
\noalign{\smallskip}
\hline
\noalign{\smallskip}
\multirow{1}{*}{1 / 0} & V-Net\scalebox{0.7}{~\cite{milletari2016v} (LB)} & 22.72 & 56.04 & 33.20 & 0* & 0.02 \\
\noalign{\smallskip}
\hline
\noalign{\smallskip}
\multirow{6}{*}{1 / 899} 
& \cellcolor{brown!20}MT \scalebox{0.7}{[NIPS'17]~\cite{tarvainen2017mean}} & \cellcolor{brown!20}33.34 & \cellcolor{brown!20}70.62 & \cellcolor{brown!20}45.09 & \cellcolor{brown!20}0.90 & \cellcolor{brown!20}16.75 \\
& \cellcolor{brown!20}\textbf{SFR$_\text{MT}$ \scalebox{0.7}{(Ours)}} & \cellcolor{brown!20}57.89 & \cellcolor{brown!20}72.09 & \cellcolor{brown!20}66.06 & \cellcolor{brown!20}71.33 & \cellcolor{brown!20}22.10 \\
& \cellcolor{brown!20}\textbf{SFR$_\text{MT}^+$ \scalebox{0.7}{(Ours)}} & \cellcolor{brown!20}65.12 & \cellcolor{brown!20}77.40 & \cellcolor{brown!20}71.35 & \cellcolor{brown!20}\textbf{77.27} & \cellcolor{brown!20}34.45 \\
& \cellcolor{teal!20}MagicNet \scalebox{0.7}{[CVPR'23]~\cite{chen2023magicnet}} & \cellcolor{teal!20}54.19 & \cellcolor{teal!20}75.26 & \cellcolor{teal!20}62.44 & \cellcolor{teal!20}48.82 & \cellcolor{teal!20}30.23 \\
& \cellcolor{teal!20}\textbf{SFR$_\text{MagicNet}$ \scalebox{0.7}{(Ours)}} & \cellcolor{teal!20}64.32 & \cellcolor{teal!20}75.04 & \cellcolor{teal!20}68.08 & \cellcolor{teal!20}76.12 & \cellcolor{teal!20}38.03 \\ 
& \cellcolor{teal!20}\textbf{SFR$_\text{MagicNet}^+$ \scalebox{0.7}{(Ours)}} & \cellcolor{teal!20}\textbf{66.16} & \cellcolor{teal!20}\textbf{79.91} & \cellcolor{teal!20}\textbf{71.63} & \cellcolor{teal!20}70.91 & \cellcolor{teal!20}\textbf{42.18} \\
\noalign{\smallskip}
\hline
\end{tabular}}}
\end{table}

%%%%%%%%%%%%%%%%%%%%% subsection %%%%%%%%%%%%%%%%%%%%%%%%
\subsection{Segmentation Results with Moderate Annotations}
\subsubsection{Single Target Segmentation}
We compare with current SOTA methods on LA dataset and BraTS dataset, including V-Net~\cite{milletari2016v}, MT~\cite{tarvainen2017mean}, UA-MT~\cite{yu2019uncertainty}, CCT~\cite{ouali2020semi}, CPS~\cite{chen2021semi}, DTC~\cite{luo2021semi}, ICT~\cite{verma2022interpolation}, CPCL~\cite{xu2022all}, URPC~\cite{luo2022semi}, BCP~\cite{bai2023bidirectional}, MCCauSSL~\cite{miao2023caussl} and ACMT~\cite{xu2023ambiguity}, and the results are shown in Table~\ref{tab:la_ssl} and Table~\ref{tab:brats_ssl}. 
We follow the same data partitioning and training strategies as in previous works~\cite{yu2019uncertainty, bai2023bidirectional, xu2023ambiguity, chen2023magicnet}, ensuring that the comparison results are fair. In all experiments, we apply data augmentation techniques such as random cropping and flipping. It could be seen that BCP~\cite{bai2023bidirectional}, CauSSL~\cite{miao2023caussl}, and ACMT~\cite{xu2023ambiguity} perform well, which may be attributed to these methods are often well-designed for small datasets, where their sophisticated architecture may yield promising results. We select the basic MT~\cite{tarvainen2017mean} network and method ACMT~\cite{xu2023ambiguity} with the current highest performance, and respectively provide pseudo-labels of adapted SAM to assist training. Our SFR framework with MT and ACMT as the retraining module is denoted as SFR$_\text{MT}$ and SFR$_\text{ACMT}$, respectively. 
The results show that our SFR framework brings substantial improvements in both MT and ACMT methods, approaching the performance of fully-supervised segmentation. Additionally, it is worth noting that SFR shows improvements in the Average Surface Distance (ASD) metric. In both datasets, our semi-supervised method SFR$_\text{ACMT}$ not only reduces the ASD but even surpasses the fully-supervised results. This improvement in ASD indicates a meaningful boost in boundary precision, which is critical for accurate medical image segmentation. 
Our SFR$^+$ framework achieves a further improvement over SFR on both datasets. In Table~\ref{tab:la_ssl}, it is clear that SFR$^+$ improves the Dice score of ACMT from 90.31\% to 91.00\%, which is only 0.14\% lower than the fully supervised result of 91.14\%.

\begin{table*}[t]
\caption{Results of BTCV dataset with scarce annotations and the evaluation metric is the Dice score (\%). Note that the full name of the organ is the same as in Table~\ref{tab:btcv_ssl}. * means all the pixels are predicted as background or another region.}
\label{tab:btcv_bsl}
\centering
\scalebox{0.92}{
\begin{tabular}{c|c|c|ccccccccccccc}
\hline
\noalign{\smallskip}
L / U & Method & AVG $\uparrow$ & Spl & R.Kid & L.Kid & Gall & Eso & Liv & Sto & Aor & IVC & Veins & Pan & RG & LG \\ 
\noalign{\smallskip}
\hline
\noalign{\smallskip}
\multirow{1}{*}{1 / 0} & V-Net\scalebox{0.7}{~\cite{milletari2016v} (LB)} & 14.84 & 52.97 & 29.59 & 2.34 & 27.63 & 0* & 75.25 & 3.33 & 0* & 0* & 0* & 1.87 & 0* & 0* \\
\noalign{\smallskip}
\hline
\noalign{\smallskip}
\multirow{6}{*}{1 / 17} & \cellcolor{brown!20}MT \scalebox{0.7}{[NIPS'17]~\cite{tarvainen2017mean}} & \cellcolor{brown!20}29.47 & \cellcolor{brown!20}67.12 & \cellcolor{brown!20}41.92 & \cellcolor{brown!20}41.01 & \cellcolor{brown!20}0.82 & \cellcolor{brown!20}0* & \cellcolor{brown!20}81.67 & \cellcolor{brown!20}3.98 & \cellcolor{brown!20}53.30 & \cellcolor{brown!20}46.81 & \cellcolor{brown!20}30.60 & \cellcolor{brown!20}15.02 & \cellcolor{brown!20}0* & \cellcolor{brown!20}0.82 \\
& \cellcolor{brown!20}\textbf{SFR$_\text{MT}$} \scalebox{0.7}{(Ours)} & \cellcolor{brown!20}39.59 \scalebox{0.8}{({\color{red}$\uparrow$10.12})} & \cellcolor{brown!20}82.87 & \cellcolor{brown!20}63.52 & \cellcolor{brown!20}\textbf{61.20} & \cellcolor{brown!20}0* & \cellcolor{brown!20}0* & \cellcolor{brown!20}87.90 & \cellcolor{brown!20}19.17 & \cellcolor{brown!20}81.68 & \cellcolor{brown!20}61.43 & \cellcolor{brown!20}33.01 & \cellcolor{brown!20}23.86 & \cellcolor{brown!20}0* & \cellcolor{brown!20}0* \\
& \cellcolor{brown!20}\textbf{SFR$_\text{MT}^+$ \scalebox{0.7}{(Ours)}} & \cellcolor{brown!20}43.51 \scalebox{0.8}{({\color{red}$\uparrow$14.04})} & \cellcolor{brown!20}\textbf{89.24} & \cellcolor{brown!20}46.81 & \cellcolor{brown!20}30.39 & \cellcolor{brown!20}\textbf{50.74} & \cellcolor{brown!20}0* & \cellcolor{brown!20}88.97 & \cellcolor{brown!20}38.99 & \cellcolor{brown!20}82.45 & \cellcolor{brown!20}55.19 & \cellcolor{brown!20}42.96 & \cellcolor{brown!20}39.90 & \cellcolor{brown!20}0* & \cellcolor{brown!20}0* \\
& \cellcolor{teal!20}MagicNet \scalebox{0.7}{[CVPR'23]~\cite{chen2023magicnet}} & \cellcolor{teal!20}40.02 & \cellcolor{teal!20}52.97 & \cellcolor{teal!20}50.22 & \cellcolor{teal!20}44.17 & \cellcolor{teal!20}10.04 & \cellcolor{teal!20}35.55 & \cellcolor{teal!20}76.02 & \cellcolor{teal!20}29.30 & \cellcolor{teal!20}56.69 & \cellcolor{teal!20}49.65 & \cellcolor{teal!20}37.59 & \cellcolor{teal!20}48.58 & \cellcolor{teal!20}16.31 & \cellcolor{teal!20}13.15 \\
& \cellcolor{teal!20}\textbf{SFR$_\text{MagicNet}$} \scalebox{0.7}{(Ours)} & \cellcolor{teal!20}53.59 \scalebox{0.8}{({\color{red}$\uparrow$13.57})} & \cellcolor{teal!20}85.01 & \cellcolor{teal!20}\textbf{67.43} & \cellcolor{teal!20}56.74 & \cellcolor{teal!20}34.77 & \cellcolor{teal!20}35.90 & \cellcolor{teal!20}86.65 & \cellcolor{teal!20}38.07 & \cellcolor{teal!20}84.58 & \cellcolor{teal!20}57.45 & \cellcolor{teal!20}56.90 & \cellcolor{teal!20}43.35 & \cellcolor{teal!20}20.94 & \cellcolor{teal!20}28.84 \\ 
& \cellcolor{teal!20}\textbf{SFR$_\text{MagicNet}^+$ \scalebox{0.7}{(Ours)}} & \cellcolor{teal!20}\textbf{58.84} \scalebox{0.8}{({\color{red}$\uparrow$18.82})} & \cellcolor{teal!20}89.02 & \cellcolor{teal!20}64.19 & \cellcolor{teal!20}55.70 & \cellcolor{teal!20}44.27 & \cellcolor{teal!20}\textbf{46.84} & \cellcolor{teal!20}\textbf{89.50} & \cellcolor{teal!20}\textbf{39.41} & \cellcolor{teal!20}\textbf{87.24} & \cellcolor{teal!20}\textbf{66.02} & \cellcolor{teal!20}\textbf{60.92} & \cellcolor{teal!20}\textbf{49.12} & \cellcolor{teal!20}\textbf{35.98} & \cellcolor{teal!20}\textbf{36.75} \\
\noalign{\smallskip}
\hline
\end{tabular}}
\end{table*}

\begin{table*}[t]
\caption{Results of MACT dataset with scarce annotations and the evaluation metric is the Dice score (\%). L.Kidey: left kidney. * means all the pixels are predicted as background or another region.}
\label{tab:mact_bsl}
\centering
\scalebox{0.92}{
\begin{tabular}{c|c|c|cccccccc}
\hline
\noalign{\smallskip}
L / U & Method & AVG $\uparrow$ & Spleen & L.Kedney & Gallbladder & Esophagus & Liver & Stomach & Pancreas & Doudenum \\ 
\noalign{\smallskip}
\hline
\noalign{\smallskip}
\multirow{1}{*}{1 / 0} & V-Net\scalebox{0.7}{~\cite{milletari2016v} (LB)} & 18.73 & 41.76 & 5.19 & 3.70 & 0.23 & 85.87 & 11.09 & 1.98 & 0.05 \\
\noalign{\smallskip}
\hline
\noalign{\smallskip}
\multirow{6}{*}{1 / 69} & \cellcolor{brown!20}MT \scalebox{0.7}{[NIPS'17]~\cite{tarvainen2017mean}} & \cellcolor{brown!20}23.09 & \cellcolor{brown!20}37.08 & \cellcolor{brown!20}29.58 & \cellcolor{brown!20}6.74 & \cellcolor{brown!20}0* & \cellcolor{brown!20}77.41 & \cellcolor{brown!20}27.65 & \cellcolor{brown!20}4.78 & \cellcolor{brown!20}1.49 \\
& \cellcolor{brown!20}\textbf{SFR$_\text{MT}$} \scalebox{0.7}{(Ours)} & \cellcolor{brown!20}36.23 \scalebox{0.8}{({\color{red}$\uparrow$13.14})} & \cellcolor{brown!20}74.41 & \cellcolor{brown!20}64.11 & \cellcolor{brown!20}7.37 & \cellcolor{brown!20}0* & \cellcolor{brown!20}91.01 & \cellcolor{brown!20}36.94 & \cellcolor{brown!20}13.37 & \cellcolor{brown!20}2.65 \\
& \cellcolor{brown!20}\textbf{SFR$_\text{MT}^+$ \scalebox{0.7}{(Ours)}} & \cellcolor{brown!20}52.54 \scalebox{0.8}{({\color{red}$\uparrow$29.45})} & \cellcolor{brown!20}76.27 & \cellcolor{brown!20}\textbf{86.97} & \cellcolor{brown!20}\textbf{65.94} & \cellcolor{brown!20}15.70 & \cellcolor{brown!20}86.85 & \cellcolor{brown!20}21.85 & \cellcolor{brown!20}35.31 & \cellcolor{brown!20}\textbf{31.42} \\
& \cellcolor{teal!20}MagicNet \scalebox{0.7}{[CVPR'23]~\cite{chen2023magicnet}} & \cellcolor{teal!20}42.90  & \cellcolor{teal!20}79.32 & \cellcolor{teal!20}62.32 & \cellcolor{teal!20}21.30 & \cellcolor{teal!20}20.87 & \cellcolor{teal!20}89.60 & \cellcolor{teal!20}44.96 & \cellcolor{teal!20}12.83 & \cellcolor{teal!20}12.01 \\
& \cellcolor{teal!20}\textbf{SFR$_\text{MagicNet}$} \scalebox{0.7}{(Ours)} & \cellcolor{teal!20}49.08 \scalebox{0.8}{({\color{red}$\uparrow$6.18})} & \cellcolor{teal!20}\textbf{89.57} & \cellcolor{teal!20}74.00 & \cellcolor{teal!20}33.21 & \cellcolor{teal!20}16.55 & \cellcolor{teal!20}\textbf{91.05} & \cellcolor{teal!20}46.91 & \cellcolor{teal!20}25.84 & \cellcolor{teal!20}15.47 \\ 
& \cellcolor{teal!20}\textbf{SFR$_\text{MagicNet}^+$ \scalebox{0.7}{(Ours)}} & \cellcolor{teal!20}\textbf{54.05} \scalebox{0.8}{({\color{red}$\uparrow$11.15})} & \cellcolor{teal!20}84.74 & \cellcolor{teal!20}85.03& \cellcolor{teal!20}36.13 & \cellcolor{teal!20}\textbf{25.01} & \cellcolor{teal!20}90.34 & \cellcolor{teal!20}\textbf{52.22} & \cellcolor{teal!20}\textbf{41.92} & \cellcolor{teal!20}17.03 \\
\noalign{\smallskip}
\hline
\end{tabular}}
\end{table*}

\begin{figure*}[t]
   \centering
   \includegraphics[width=1.0\linewidth]{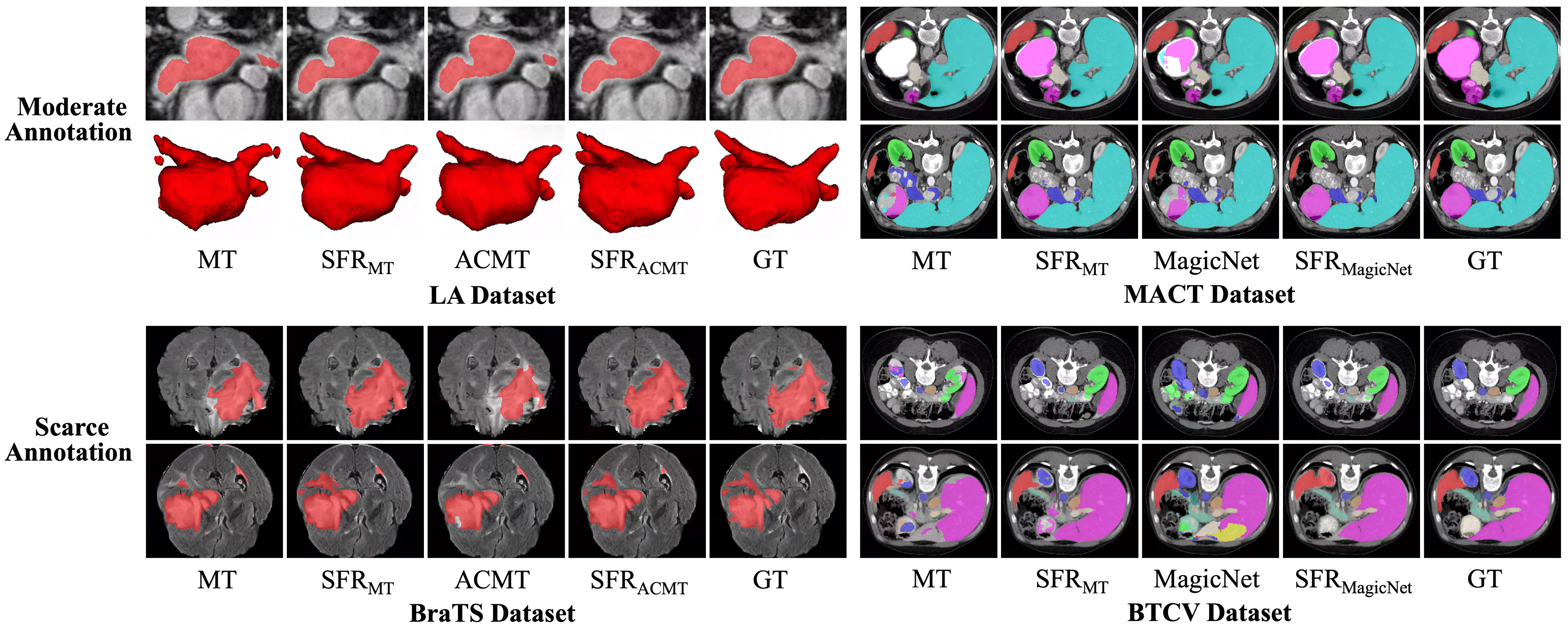}
   \caption{Segmentation results on LA and MACT datasets with moderate annotation, and on BraTS and BTCV datasets with scarce annotation.}
   \label{fig: all_visual}
\end{figure*}

\subsubsection{Multiple Target Segmentation}
Table~\ref{tab:abdomen_ssl}, Table~\ref{tab:btcv_ssl}, and Table~\ref{tab:mact_ssl} show the results on AbdomenCT-1K, BTCV, and MACV datasets, respectively. We compare our method with V-Net~\cite{milletari2016v}, MT~\cite{tarvainen2017mean}, UA-MT~\cite{yu2019uncertainty}, CPS~\cite{chen2021semi}, ICT~\cite{verma2022interpolation}, SS-Net~\cite{wu2022exploring}, SLC-Net~\cite{liu2022semi, liu2023semi}, and MagicNet~\cite{chen2023magicnet}.
We employ SFR framework to guide the classical MT~\cite{tarvainen2017mean} and the SOTA MagicNet~\cite{chen2023magicnet} methods, \ie, SFR$_\text{MT}$ and SFR$_\text{MagicNet}$.
By leveraging the SFR framework, both methods achieve significant performance improvements. On the BTCV dataset, SFR$_\text{MagicNet}$ even surpasses the results obtained with fully-supervised learning. 
For the AbdomenCT-1K dataset, it could be observed that SFR$_\text{MT}$ improves performance across all four organs, boosting the average Dice score by 13.57\% (88.73\% vs. 75.16\%) with 20\% labeled data.
In addition, our SFR$^+$ framework demonstrates improvements across the three different datasets for moderate annotation scenarios, highlighting its robustness and effectiveness in segmentation.

The single target segmentation examples on LA dataset with 16 labeled data and multiple target segmentation examples on MACT dataset with 14 labeled data are shown in Fig.~\ref{fig: all_visual}. Our SFR$_\text{MT}$ and SFR$_\text{ACMT}$ predictions align more accurately with ground truth masks, further validating the effectiveness. Our framework demonstrates its ability to adapt different data.

%%%%%%%%%%%%%%%%%%%%% subsection %%%%%%%%%%%%%%%%%%%%%%%%
\subsection{Segmentation Results with Scarce Annotations}
\subsubsection{Single Target Segmentation}
We evaluate the impact of SFR with scarce annotations on the LA and BraTS datasets.
In Table~\ref{tab:la_bsl}, when only one labeled volume is available, supervised learning performs poorly, achieving only 17.99\% accuracy. SFR takes advantage of medical image stitching and SAM's powerful feature capture capabilities to guide MT to increase from 29.68\% to 74.40\%. 
The results of the BraTS are shown in Table~\ref{tab:brats_bsl} and SFR helps MT improve by 15.06\% and MagicNet by 15.15\%. In addition, one phenomenon observed is that some semi-supervised methods perform worse than the lower bound. This could be attributed to the incomplete similarity of the data distribution between labeled and unlabeled samples, and the inaccurate pseudo-labels from the unlabeled samples have a negative impact on the model's learning process.
Furthermore, our SFR$^+$ framework exploits a confidence-based selective strategy and achieves better results than SFR on both MT and ACMT methods.

We recognize that there is a large performance difference between MT with moderate and scarce annotations on the LA dataset compared to the BraTS dataset. To better understand this performance improvement with the increase of labeled samples, we perform kernel dense estimation on different settings, as shown in Fig.~\ref{fig:kde_la_brats}. In the LA dataset, the results show that with scarce annotation, the distribution gap between labeled and unlabeled data is quite large. This may cause MT to struggle in feature extraction because it relies on this labeled set to learn the overall data distribution. As the number of labeled samples increases, the distribution gap between labeled and unlabeled data becomes small, so the effect of MT improves rapidly. For BraTS dataset, the difference between the feature distributions of labeled and unlabeled data is not as pronounced as in LA dataset with scarce annotation, which may lead to a better effect on BraTS than on LA dataset.

\subsubsection{Multiple Target Segmentation}
The results of scarce annotations on three multi-organ datasets are presented in Table~\ref{tab:abdomen_bsl}, Table~\ref{tab:btcv_bsl}, and Table~\ref{tab:mact_bsl}.
On BTCV dataset with only one labeled sample, the model struggles to identify esophagus, aorta, inferior vena cava, portal and splenic veins, left and right adrenal glands, with Dice scores below 5\% for left kidney, stomach, and pancreas.
Semi-supervised methods with unlabeled samples alleviate the poor performance of most organs.
Our SFR framework achieves improvements of 10.12\% and 13.57\% on MT and MagicNet models, respectively, particularly with a nearly 30\% improvement on aorta region.
Similarly, on MACT dataset, SFR helps identify challenging classes, with significant improvements of 37.33\% for spleen and 34.53\% for right kidney on the Mean Teacher model.
Also, SFR$_\text{MagicNet}$ achieves a performance improvement of up to 10.13\% (64.32\% vs. 66.16\%) Dice score on the AbdomenCT-1K dataset with one annotation.
Our SFR$^+$ framework further improves performance, achieving more than 10\% improvement in the Dice metric on all three datasets.

Fig.~\ref{fig: all_visual} shows the segmentation examples of single target dataset BraTS and multiple target dataset BTCV with scarce annotation. 
The effectiveness of our proposed framework can be shown in some challenging examples. 
In these examples, the baselines (\ie, MT~\cite{tarvainen2017mean}, ACMT~\cite{xu2023ambiguity} and MagicNet~\cite{chen2023magicnet}) tend to generate false predictions or incomplete structures, whereas the introduction of SFR framework can mitigate these problems and obtain a more plausible segmentation result.
Our framework demonstrates its ability to adapt flexibly to different task requirements. This versatility highlights the potential of our framework to handle a diverse range of medical imaging tasks with varying levels of annotation availability.

\begin{table}[t]
\caption{Comparison of fine-tuning and re-training segmentation results of Dice on four datasets. ``\# L" is the number of labeled data. "M" means moderate and "S" means scarce. }
\label{tab:retraining_acc}
\centering
\scalebox{0.88}{
\begin{tabular}{c|c|c|cccc}
\hline
\noalign{\smallskip}
\multirow{2}{*}{\#L} & \multirow{2}{*}{Method} & \multirow{2}{*}{Params$\downarrow$} & \multicolumn{4}{c}{Dice$\uparrow$} \\
& & & LA & BraTS & BTCV & MACT \\
\noalign{\smallskip}
\hline
\noalign{\smallskip}
\multirow{7}{*}{M} & SFR$_\text{FT}$ & 91M & 90.39 & 84.94 & 65.26 & 79.59 \\
\noalign{\smallskip}
\cline{2-7}
\noalign{\smallskip}
& \multirow{2}{*}{SFR$_\text{MT}$} & \multirow{2}{*}{10M} & 90.86 & 85.19 & 70.09 & 82.50 \\
& & & ({\color{red}$\uparrow$0.47}) & ({\color{red}$\uparrow$0.25}) & ({\color{red}$\uparrow$4.83}) & ({\color{red}$\uparrow$2.91}) \\
\noalign{\smallskip}
\cline{2-7}
\noalign{\smallskip}
& \multirow{2}{*}{SFR$_\text{ACMT/MagicNet}$} & \multirow{2}{*}{10/18M} & \textbf{90.95} & \textbf{85.81} & \textbf{77.06} & \textbf{82.87} \\
& & & ({\color{red}$\uparrow$0.56}) & ({\color{red}$\uparrow$0.87}) & ({\color{red}$\uparrow$11.80}) & ({\color{red}$\uparrow$3.28}) \\
\noalign{\smallskip}
\hline
\noalign{\smallskip}
\multirow{7}{*}{S} & SFR$_\text{FT}$ & 91M & 74.78 & 78.14 & 40.98 & 35.13 \\
\noalign{\smallskip}
\cline{2-7}
\noalign{\smallskip}
& \multirow{2}{*}{SFR$_\text{MT}$} & \multirow{2}{*}{10M} & 74.40 & \textbf{78.58} & 39.59 & 36.23 \\
& & & ({\color{blue}$\downarrow$0.38}) & ({\color{red}$\uparrow$0.44}) & ({\color{blue}$\downarrow$1.39}) & ({\color{red}$\uparrow$1.10}) \\
\noalign{\smallskip}
\cline{2-7}
\noalign{\smallskip}
& \multirow{2}{*}{SFR$_\text{ACMT/MagicNet}$} & \multirow{2}{*}{10/18M} & \textbf{76.76} & 78.47 & \textbf{53.59} & \textbf{49.08} \\
& & & ({\color{red}$\uparrow$1.98}) & ({\color{red}$\uparrow$0.33}) & ({\color{red}$\uparrow$12.61}) & ({\color{red}$\uparrow$13.95}) \\
\noalign{\smallskip}
\hline
\end{tabular}}
\end{table}

%%%%%%%%%%%%%%%%%%%%% subsection %%%%%%%%%%%%%%%%%%%%%%%%
\subsection{Effectiveness of Re-training Module}
We report the results of fine-tuning step omitting re-training in all datasets in Table~\ref{tab:retraining_acc}, and compare the model parameters and performance after fine-tuning and re-training steps.  
By comparing fine-tuning results with re-training results in more detail, we observe that there is a slight performance decrease in SFR$_\text{MT}$ with scarce labels on LA and BTCV datasets, but overall there is a clear improvement in effect.
On the other hand, considering the practical deployment requirements, re-training model has a smaller number of parameters than fine-tuning model. Our fine-tuning module SFR$_\text{FT}$ has the same parameter size as the foundation model (SAM) and the re-training module reduces the parameter scale to the mainstream segmenters.
The results validate that the re-training module contributes to an overall performance boost for semi-supervised segmentation. 

\begin{figure}[t]
  \centering
   \includegraphics[width=0.95\linewidth]{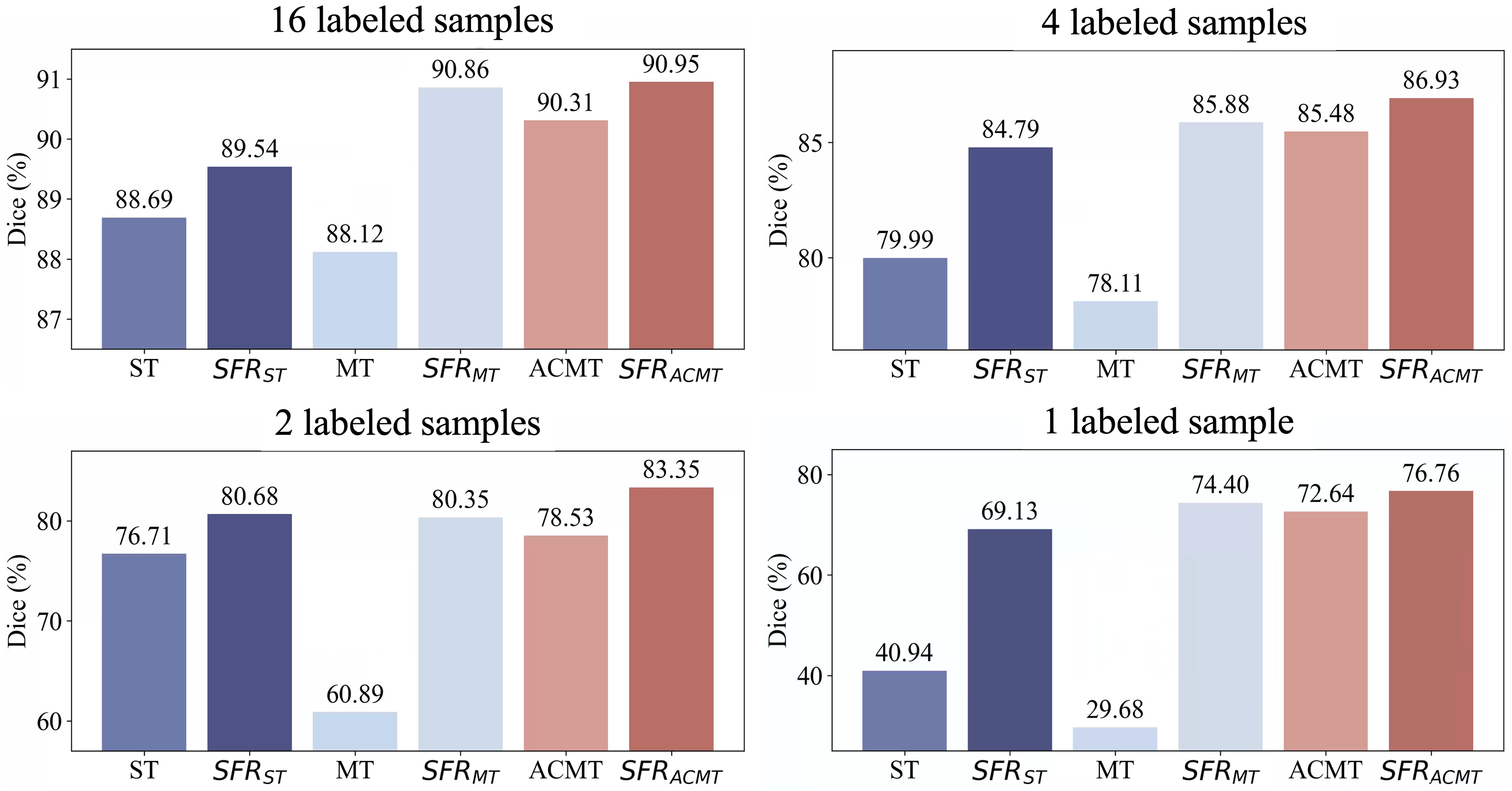}
   \caption{Results of SFR on self-training (ST in short), MT, and ACMT with different numbers of labeled samples.}
   \label{fig: ablation_ssl}
\end{figure}

%%%%%%%%%%%%%%%%%%%%% subsection %%%%%%%%%%%%%%%%%%%%%%%%
\subsection{Further Analysis}
\subsubsection{Analysis of Initial Pseudo-labels}
We compare our fine-tuning module of SFR framework with four foundation model fine-tuning methods, including MedSAM-v1~\cite{ma2023segment}, MedSAM-v2~\cite{ma2023segment}, SAMed~\cite{zhang2023customized} and SAMUS~\cite{lin2023samus}.
MedSAM adopts a sub-parts fine-tuning approach and SAMed is a LoRA-based adapter tuning method, which maintains the same parameter scale as the original SAM model. 
SAMUS introduces a parallel CNN branch and thus has more parameters. 
As shown in Fig.~\ref{fig: intro}, it demonstrates that our stitching fine-tuning strategy (denoted as SFR$_\text{FT}$) achieves a significant performance improvement without introducing additional parameters, and therefore could provide more reliable pseudo-labels for subsequent retraining module. 

\subsubsection{Analysis of Pseudo-label Error Propagation}
In our work, we use fine-tuned foundation model to generate pseudo-labels, and the errors in pseudo-labels on unlabeled data could propagate and impact the model’s performance. We have considered the potential challenge and implemented some measures to address it effectively. Firstly, within our SFR framework, the semi-supervised retraining module does not solely rely on the pseudo-labels generated by the fine-tuned SAM models. Instead, we incorporate a semi-supervised consistency loss that encourages stable predictions across various perturbations of the input. This helps the model learn consistent and robust features, reducing its sensitivity to minor inaccuracies or noise in the pseudo-labels. Secondly, in the extended version of our framework, SFR$^+$, we introduce confidence estimation to better handle each unlabeled sample. This approach ensures that the model’s learning process is guided primarily by more reliable pseudo-labels, reducing the impact of erroneous labels on the overall performance.

\subsubsection{Compatibility under Different Number of Labeled Data}
With its plug-and-play nature, our re-training module could apply different SSL methods. 
We conduct re-training experiments on self-training, MT, and ACMT methods with different numbers of labeled samples in Fig.~\ref{fig: ablation_ssl}. 
Compared to the baseline, our SFR yields excellent segmentation performance with 1, 2, 4, or 16 labeled samples.
Especially in the context of extreme annotation conditions, \ie, 1 labeled sample, our framework produces impressive improvements.

\begin{table}[t]
\renewcommand\arraystretch{0.9}
\caption{Comparison of 3D-based fine-tuning method on LA dataset. }
\label{tab:3dsam}
\centering
\scalebox{0.95}{
\begin{tabular}{c|c|cccc}
\hline
\noalign{\smallskip}
Method & Params$\downarrow$ & Dice$\uparrow$ & Jaccard$\uparrow$ & ASD$\downarrow$ & 95HD$\downarrow$ \\
\noalign{\smallskip}
\hline
\noalign{\smallskip}
3DSAM-Adapter\scalebox{0.7}{~\cite{gong20233dsam}} & 114M & 82.39 & 70.85 & 3.74 & 15.49 \\
SFR$_\text{FT}$ & 91M & \textbf{90.39} & \textbf{82.54} & \textbf{1.83} & \textbf{6.71} \\
\noalign{\smallskip}
\hline
\end{tabular}}
\end{table}

\begin{table}[t]
\caption{Comparison with SAM-Med3D on LA dataset. }
\label{tab:medsam3d}
\centering
\scalebox{0.85}{
\textcolor{black}{
\begin{tabular}{c|cc|c|c|c|c}
\hline
\noalign{\smallskip}
Method & Prompt & Number & Dice$\uparrow$ & Jaccard$\uparrow$ & ASD$\downarrow$ & 95HD$\downarrow$ \\
\noalign{\smallskip}
\hline
\noalign{\smallskip}
\multirow{3}{*}{SAM-Med3D\scalebox{0.7}{~\cite{wang2023sammed3d}}} & click & 2 & 57.33 & 42.31 & 8.58 & 28.88 \\ 
& click & 5 & 65.12 & 50.29 & 10.65 & 31.17 \\ 
& click & 10 & 67.17 & 52.54 & 11.05 & 33.37 \\ 
\noalign{\smallskip}
\hline
\noalign{\smallskip}
\multicolumn{3}{c|}{SFR (Ours) - scarce annotations (1)} & 76.76 & 64.65 & 8.54 & 28.07 \\ 
\multicolumn{3}{c|}{SFR (Ours) - moderate annotations (20\%)} & \textbf{90.95} & \textbf{83.47} & \textbf{1.43} & \textbf{6.11} \\ 
\noalign{\smallskip}
\hline
\end{tabular}}}
\end{table}

\subsubsection{Compared with 3D-based SAM methods}
We compare the results of our fine-tuning SFR$_\text{FT}$ and the existing 3D-based SAM fine-tuning method (3DSAM-Adapter) in Table~\ref{tab:3dsam}, which reveals that our method achieves better accuracy without additional computational overhead.
Our method outperforms the 3DSAM-Adapter by 8\%. A possible explanation for this performance difference lies in the size of the training parameters. For small-scale medical imaging datasets, particularly in scenarios where training data is limited, a model with a larger number of parameters, which has high complexity, may be prone to overfitting. In contrast, our method effectively utilizes 2D slice stitching input with the low-rank-based fine-tuning strategy. It allows the model to learn spatial relationships while managing computational complexity, which results in better generalization and segmentation performance.

In addition, we conduct a comparison with the SAM-Med3D method~\cite{wang2023sammed3d}, which modifies the SAM paradigm to 3D architecture with training from scratch on 3D medical image datasets. We compare the results of our framework with the prompt-based SAM-Med3D method on the same test data in Table~\ref{tab:medsam3d}. Notably, on the LA dataset, our SFR framework, using just a single annotation, outperforms SAM-Med3D with 2, 5, or 10-point prompts per sample.

\subsubsection{Compared with SAM 2-based methods}
We evaluated both SAM 2~\cite{ravi2024sam} and Medical SAM 2~\cite{zhu2024medical} across multiple datasets, including LA, BraTS, BTCV, and MACT, under different prompt conditions (2, 5, and 10 clicks), as shown in Table~\ref{tab: sam2}. While these models perform well in certain datasets, such as BTCV and MACT, our SFR framework demonstrates significantly superior performance when moderate annotations are provided. Moreover, for the LA and BraTS datasets, the Dice scores of our method, even with scarce annotations (1 annotation), are noticeably higher than the results obtained by both SAM 2 and Medical SAM 2 under any prompt frequency condition. This further underscores the robustness of our approach, particularly when dealing with limited annotations.

\begin{table}[t]
\caption{Comparison with foundation models evaluated by Dice score. }
\label{tab: sam2}
\centering
\scalebox{0.85}{
\textcolor{black}{
\begin{tabular}{c|cc|c|c|c|c}
\hline
\noalign{\smallskip}
Method & Prompt & Frequency & LA & BraTS & BTCV & MACT \\
\noalign{\smallskip}
\hline
\noalign{\smallskip}
\multirow{3}{*}{SAM 2\scalebox{0.7}{~\cite{ravi2024sam}}} & click & 2 & 31.13 & 25.68 & 55.64 & 77.09 \\
& click & 5 & 30.73 & 25.08 & 54.07 & 73.12 \\ 
& click & 10 & 29.41 & 23.88 & 53.48 & 72.40 \\ 
\noalign{\smallskip}
\hline
\noalign{\smallskip}
\multirow{3}{*}{Medical SAM 2\scalebox{0.7}{~\cite{zhu2024medical}}} & click & 2 & 31.64 & 28.91 & 59.47 & 78.77 \\ 
& click & 5 & 33.56 & 28.06 & 56.78 & 76.67 \\ 
& click & 10 & 30.38 & 25.85 & 56.27 & 75.42 \\ 
\noalign{\smallskip}
\hline
\noalign{\smallskip}
\multicolumn{3}{c|}{SFR (Ours) - scarce annotations} & 76.76 & 78.58 & 53.59 & 49.08 \\ 
\multicolumn{3}{c|}{SFR (Ours) - moderate annotations} & \textbf{90.95} & \textbf{85.81} & \textbf{77.06} & \textbf{82.87} \\ 
\noalign{\smallskip}
\hline
\end{tabular}}}
\end{table}

\subsubsection{Compared with Concatenating along Channel Dimension}
By concatenating three slices along the channel dimension to create a pseudo-RGB image, SAM could potentially segment this combined representation. On the other hand, as the SAM model is pre-trained on natural color images, it might interpret the concatenated channels more as color information than as meaningful anatomical context. This presents an intriguing yet challenging question, that is worth exploring further in future research to better evaluate its applicability and uncover potential benefits.

\subsubsection{Spatial Structure of Medical Images} 
During fine-tuning, regarding the distribution gap and dimensionality mismatch, we believe the global scope of all the consecutive slices could be a bridge to link the SAM (designed for 2D natural images) and 3D medical images. As we demonstrated in Section~\ref{sec:concat}, our stitching method outperforms the listed alternatives, indicating that in the fine-tuning step, SAM benefits from accurately locating the regions to segment due to its generalization ability in various segmentation tasks.
During re-training, we still use 3D structure-based SSL methods, to retain spatial information. 
In a nutshell, SAM in fine-tuning and SSL in re-training separately treat 3D information in different ways: SAM to solve stitched large-sized 2D slices, and SSL to further segment the 3D volumes.

\section{Conclusion}
\label{sec:conclusion}
In this work, we present the SFR framework, which consists of the stitching, fine-tuning, and re-training modules, to achieve higher improvements in semi-supervised segmentation tasks by leveraging the foundation model.
The stitching module copes with the resolution difference between medical and natural images, and fine-tuning module provides reliable initial pseudo-labels for re-training module.
Our framework maintains the same parameter size as the mainstream segmenter, \eg, V-Net~\cite{milletari2016v}, and could be compatible with most popular SSL methods, \eg, Mean Teacher~\cite{tarvainen2017mean}. 
In addition, we develop the SFR$^+$, which further enhances the framework by introducing confidence estimation and selective training strategy. 
Extensive experiments demonstrate that the SFR and SFR$^+$ frameworks improve performance remarkably in both moderate and scarce annotation scenarios.

\bibliographystyle{IEEEtran}
\bibliography{ref}

\end{document}